\documentclass[dvipsnames,format=sigconf]{acmart} 

\AtBeginDocument{%
  \providecommand\BibTeX{{%
    \normalfont B\kern-0.5em{\scshape i\kern-0.25em b}\kern-0.8em\TeX}}}

\copyrightyear{2023} \acmYear{2023} \setcopyright{rightsretained} 

\usepackage{multirow}

\begin{document}

\title[Computational and Exploratory Landscape Analysis of the GKLS Generator]{Computational and Exploratory Landscape Analysis\\ of the GKLS Generator}

\author{Jakub Kudela}
\orcid{0000-0002-4372-2105}
\affiliation{%
  \institution{Institute of Automation and Computer Science\\ Brno University of Technology}
  \city{Brno}
  \country{Czech Republic}
}
\email{Jakub.Kudela@vutbr.cz}

\author{Martin Juricek}
\orcid{0000-0002-7943-8659}
\affiliation{%
  \institution{Institute of Automation and Computer Science\\ Brno University of Technology}
  \city{Brno}
  \country{Czech Republic}
}
\email{200543@vutbr.cz}








\renewcommand{\shortauthors}{Kudela and Juricek}

\begin{abstract}
  The GKLS generator is one of the most used testbeds for benchmarking global optimization algorithms. In this paper, we conduct both a computational analysis and the Exploratory Landscape Analysis (ELA) of the GKLS generator. We utilize both canonically used and newly generated classes of GKLS-generated problems and show their use in benchmarking three state-of-the-art methods (from evolutionary and deterministic communities) in dimensions 5 and 10. We show that the GKLS generator produces ``needle in a haystack'' type problems that become extremely difficult to optimize in higher dimensions. Furthermore, we conduct the ELA on the GKLS generator and then compare it to the ELA of two other widely used benchmark sets (BBOB and CEC 2014), and discuss the meaningfulness of the results.
\end{abstract}






\maketitle

\section{Introduction}
Solving real-world black-box optimization problems is a very challenging task, even when one has domain knowledge and experience \cite{long2022bbob}, especially in problems with expensive function evaluation that require simulation runs \cite{kudela2022recent}. The development, utilization, and comparison of optimization methods on such real-world problems are usually prohibitively expensive. For such tasks, benchmarking optimization methods on artificially constructed testbeds become pivotal, with the expectation that the behavior of the methods on these benchmark sets translates well into real-world problems. 

Over the years, various benchmark suites have been proposed, in which different global function properties are represented, such as multi-modality, separability, ill-conditioning, and various other types of global structures. In the evolutionary computation community, the two most utilized benchmark sets are the Black-Box Optimization Benchmarking (BBOB) suite \cite{hansen12} which is now part of the COCO platform \cite{hansen2021coco}, and the suites that were presented at the Congress on Evolutionary Computation (CEC) competitions (which started in 2005 and continue to this day) \cite{kudela2022critical}. As was shown in \cite{garden2014analysis}, the characteristics of the functions used in these two benchmarks are quite different. The CEC benchmarks are constructed by using similar subfunctions, which possibly gives an advantage to methods that perform well on these fewer subfunctions. It was also found that the CEC functions share more similarities among themselves than with those found in the BBOB \cite{garden2014analysis}.

In the global (deterministic) optimization community, one of the most popular benchmark sets is the one produced by the GKLS generator \cite{gaviano2003algorithm}. The GKLS generator constructs classes of test functions (either non-differentiable, differentiable, or twice-differentiable) for multi-modal, multi-dimensional box-constrained global optimization. The advantage of the GKLS generator is that for each generated problem, the location and function value of its local and global minima are known. Although the GKLS generator can be used to create various types of problems (based on input parameters), there are 8 classes of problems (2 for dimensions 2, 3, 4, and 5) each containing 100 functions that are generally used \cite{sergeyev2006global,sergeyev2018efficiency,paulavivcius2020globally,stripinis2022extensive}. The GKLS generator was also recently used for the construction of nonlinear model predictive control \cite{xu2023n} and general-constrained \cite{sergeyev2021generator} test problems.

In order to quantify the low-level properties of optimization problems, various features of the landscape can be computed \cite{long2022bbob}. Such analysis falls under the field of Exploratory Landscape Analysis (ELA) \cite{mersmann2010benchmarking}. The landscape features try to approximate different aspects of the optimization problem, such as its modality, separability, or whether or not the problem has plateaus. The most notable uses of ELA are in the visualization of the problem space of various optimization benchmark problem sets \cite{vskvorc2020understanding}, and in automated algorithm selection \cite{kerschke2019automated}.

In this paper, we conduct computational analysis and ELA of the GKLS-generated problems. The rest of the paper is structured as follows. Section 2 briefly describes the GKLS generator. In Section 3, we conduct the computational analysis of benchmarking three state-of-the-art methods on three classes of GKLS-generated suits (two canonical and one newly proposed) in dimensions 5 and 10. In Section 4 the ELA of the GKLS-generated suits is presented, along with the comparison of ELA of the BBOB and CEC 2014 suits. Finally, conclusions are drawn in Section 5.

\begin{figure*}
    \centering
    \begin{tabular}{ccc}
       \includegraphics[width=0.3\linewidth]{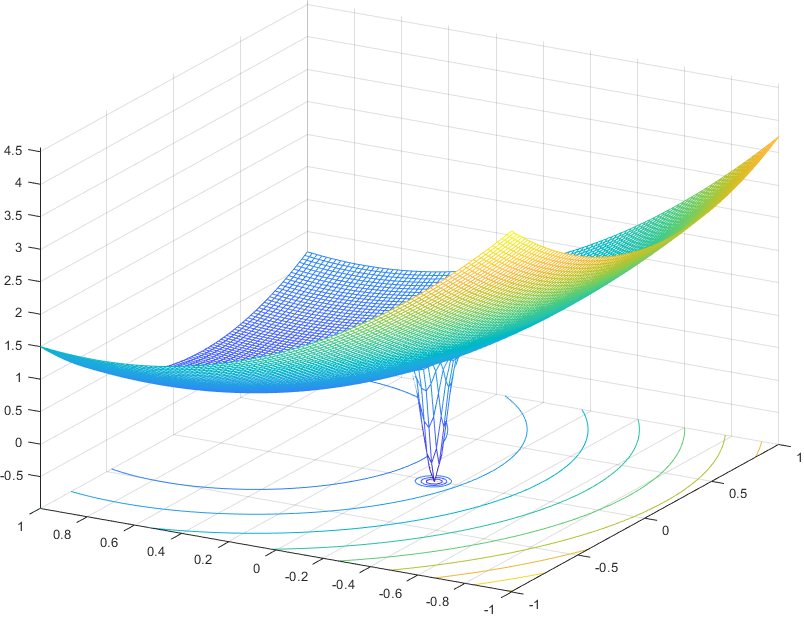}  &  \includegraphics[width=0.3\linewidth]{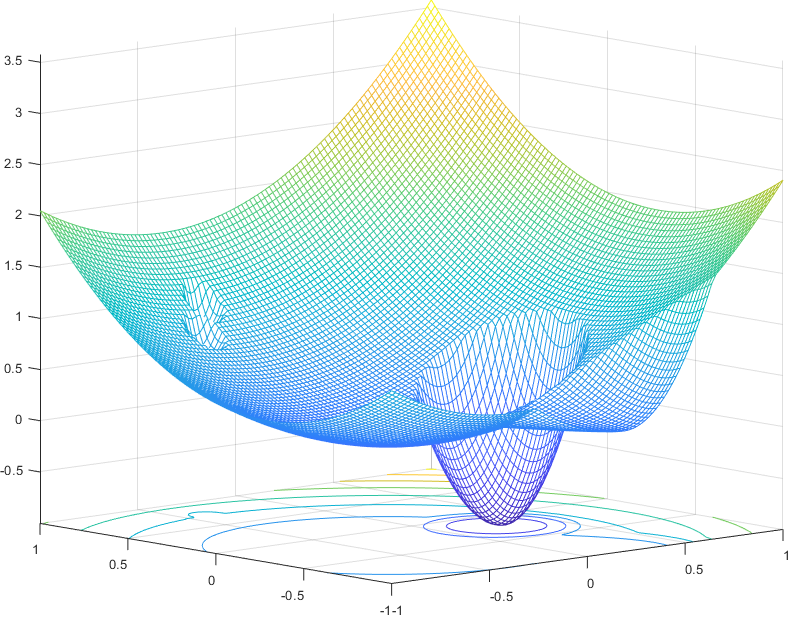} & \includegraphics[width=0.3\linewidth]{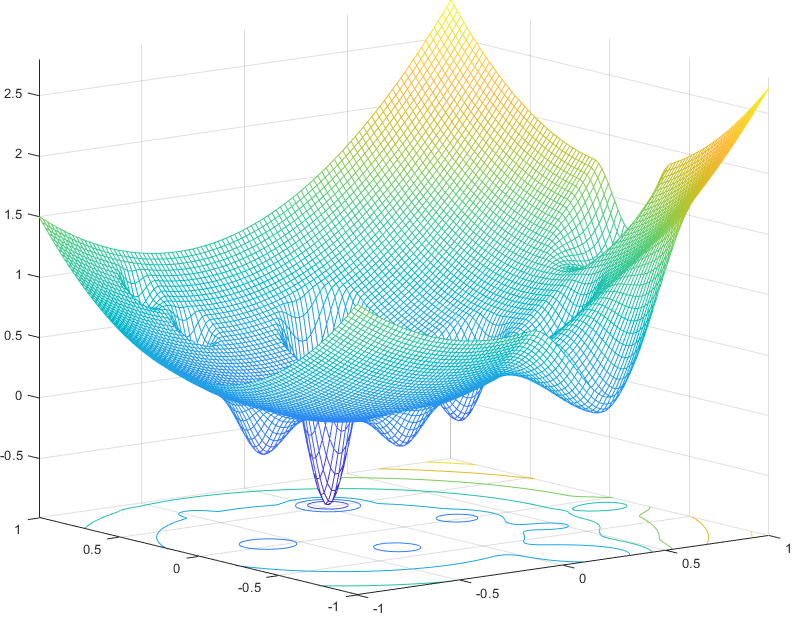}\\
        $h = 2$, type D, $d = 0.9$, $r = 0.1$ & $h = 5$, type ND, $d = 0.66$, $r = 0.5$ & $h = 10$, type D, $d = 0.66$, $r = 0.3$ \\
        \includegraphics[width=0.3\linewidth]{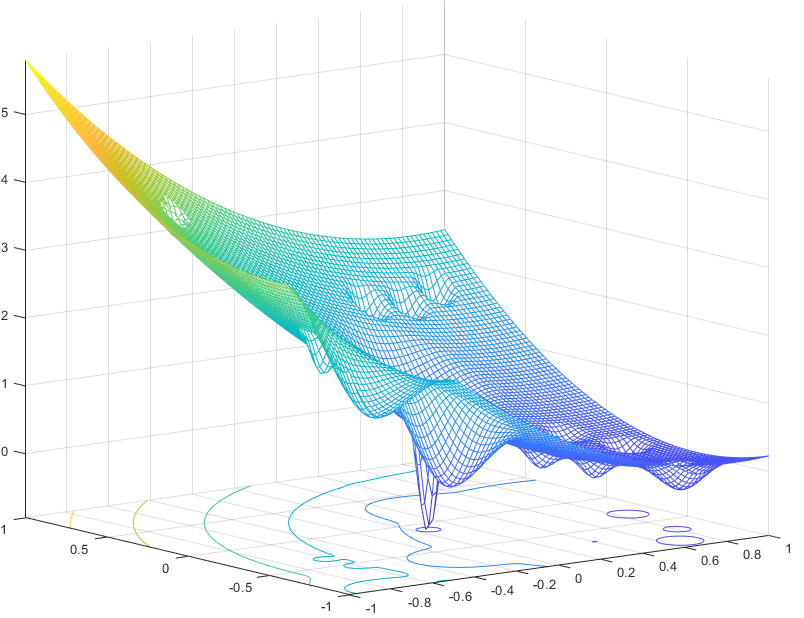}  &  \includegraphics[width=0.3\linewidth]{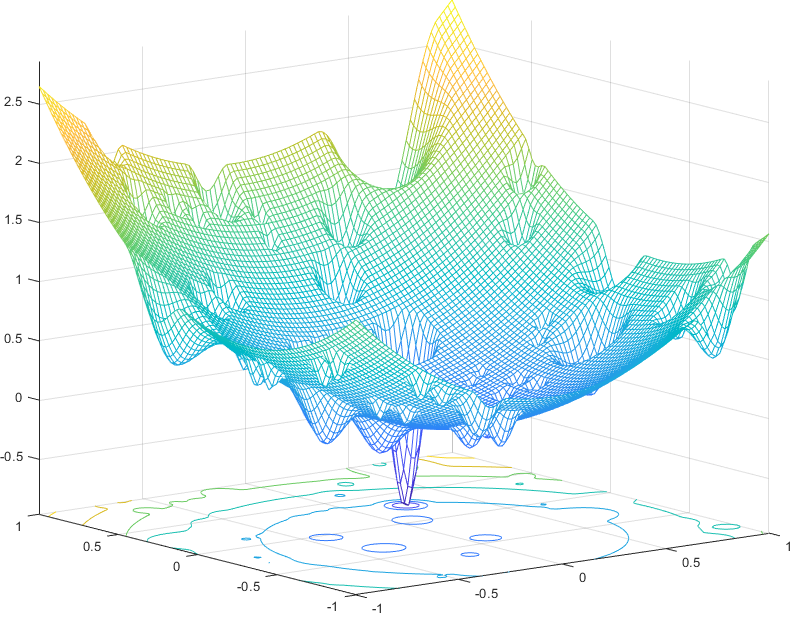} & \includegraphics[width=0.3\linewidth]{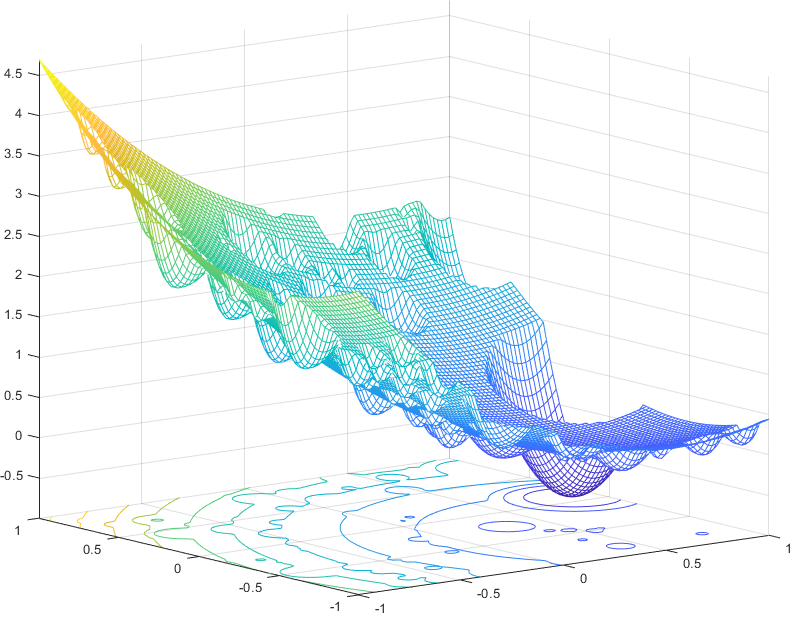}\\
        $h = 20$, type D, $d = 0.9$, $r = 0.1$ & $h = 50$, type D, $d = 0.66$, $r = 0.2$ & $h = 100$, type ND, $d = 0.66$, $r = 0.5$
    \end{tabular}
    \caption{Functions generated by the GKLS in dimension $D = 2$ ($f^* = -1$ for all functions).}
    \label{fig:f_plots}
\end{figure*}

\section{GKLS generator}
In the GKLS generator, a prespecified number of test problems (a class of problems) is constructed by defining a convex quadratic function (a paraboloid) which is systematically distorted by polynomials in order to produce local (and one global) minima. The input parameters for this construction are the following:
type of the problem (ND: non-differentiable, D: differentiable, D2: twice-differentiable) problem dimension ($D$), number of local minima ($h$), the value of the global minimum ($f^*$), radius ($r$) of the attraction region of the global minimizer, and the distance ($d$) from the global minimizer to the vertex of the quadratic function. All problems are constructed on $[-1,1]^D$.

A visualization of different functions that can be generated by the GKLS and the effect of different parameter choices is shown in Figure \ref{fig:f_plots}. Several interesting observations can be made regarding the generated functions. Firstly, they are all relatively well-conditioned. The local minimum of the ``big'' paraboloid is always in the domain and has a function value of 0. The ``attraction regions'' of the different local minima do not overlap (this is by design) - this also means that they become more shallow when their number increases.

The eight most used classes (each with 100 functions) of GKLS-generated problems are shown in Table \ref{tab:gkls}. An interesting thing to note is that these classes are quite similar, with type, $f^*$, and $h$ parameters being the same. Another feature of the GKLS generator is that it has a built-in pseudorandom number generator. If the same parameters are given as an input to GKLS an identical class of functions will be produced. However, the pseudorandom number generator (or, more precisely, its seed) does not depend on all input parameters but is mainly dependent on $h$. This has some interesting effects. For instance, two problems in different classes in the same dimension and with the same number (i.e., problem number 1 in class 1 and problem number 1 in class 2), and with the same value of $h$ will have the same location of the local minimum of the ``big'' paraboloid. If the classes also share the same $d$, the problems with the same number will also have the same location of the global minimum. This issue can be seen in problem classes 1 and 2, and 7 and 8.

\begin{table}[]
\caption{Most used GKLS test classes.}
\label{tab:gkls}
\begin{tabular}{cccccccc}
Class &  ``Difficulty'' & Type & $D$ & $f^*$  & $d$    & $r$    & $h$  \\ \hline 
1     & simple   & D&2 & -1 & 0.90 & 0.20 & 10 \\
2     & hard     &D& 2 & -1 & 0.90 & 0.10 & 10 \\ \hline 
3     &  simple   &D& 3 & -1 & 0.66 & 0.20 & 10 \\
4     &  hard     &D& 3 & -1 & 0.90 & 0.20 & 10 \\ \hline
5     &  simple   &D& 4 & -1 & 0.66 & 0.20 & 10 \\
6     &hard     &D& 4 & -1 & 0.90 & 0.20 & 10 \\ \hline 
7     &  simple   &D& 5 & -1 & 0.66 & 0.30 & 10 \\
8     & hard     &D& 5 & -1 & 0.66 & 0.20 & 10 \\
\end{tabular}
\end{table}

\begin{figure*}
    \centering
    \begin{tabular}{ccc}
       \includegraphics[width=0.3\linewidth]{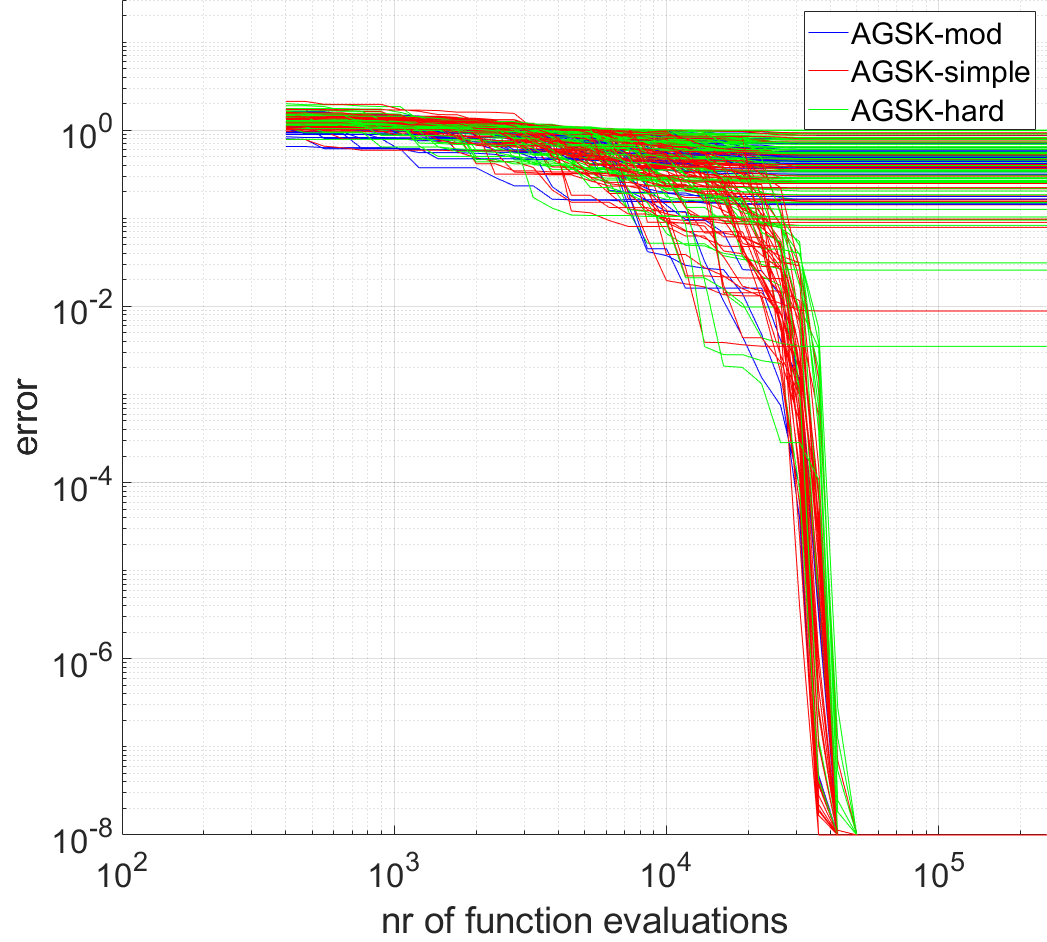}  &  \includegraphics[width=0.3\linewidth]{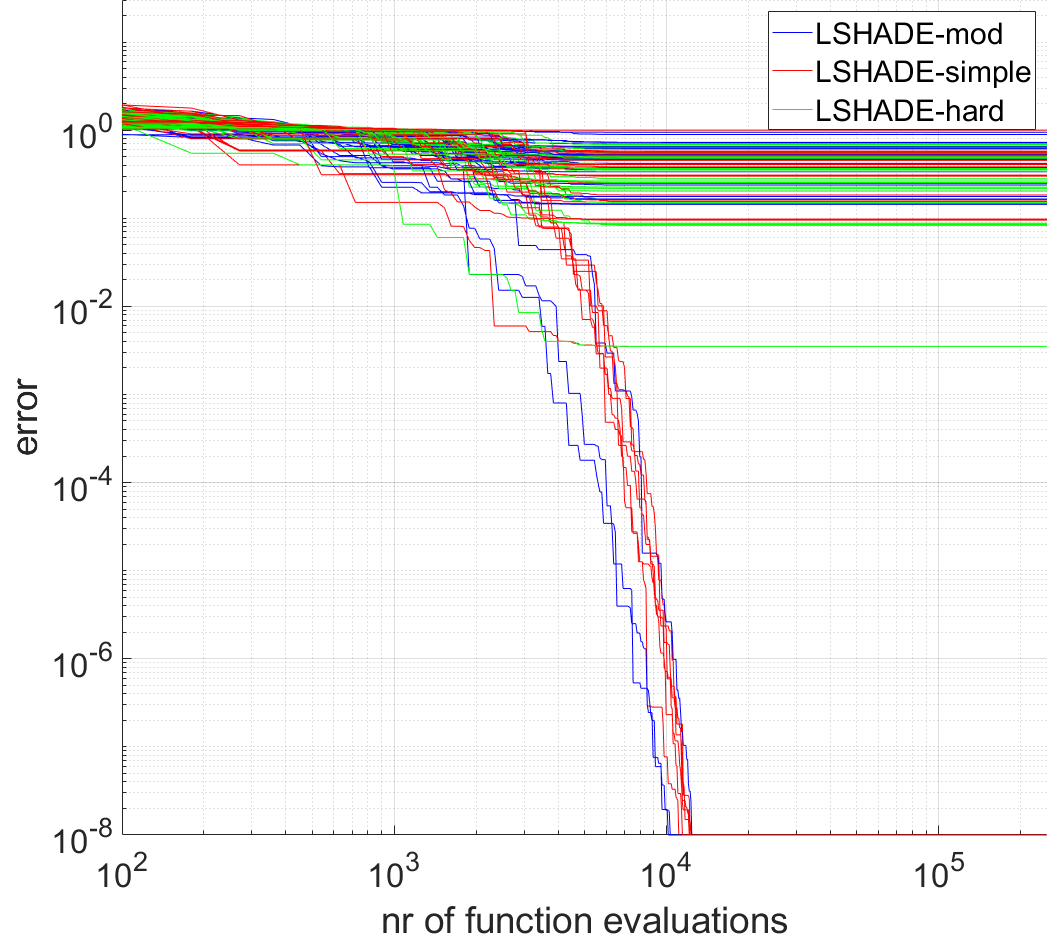} & \includegraphics[width=0.3\linewidth]{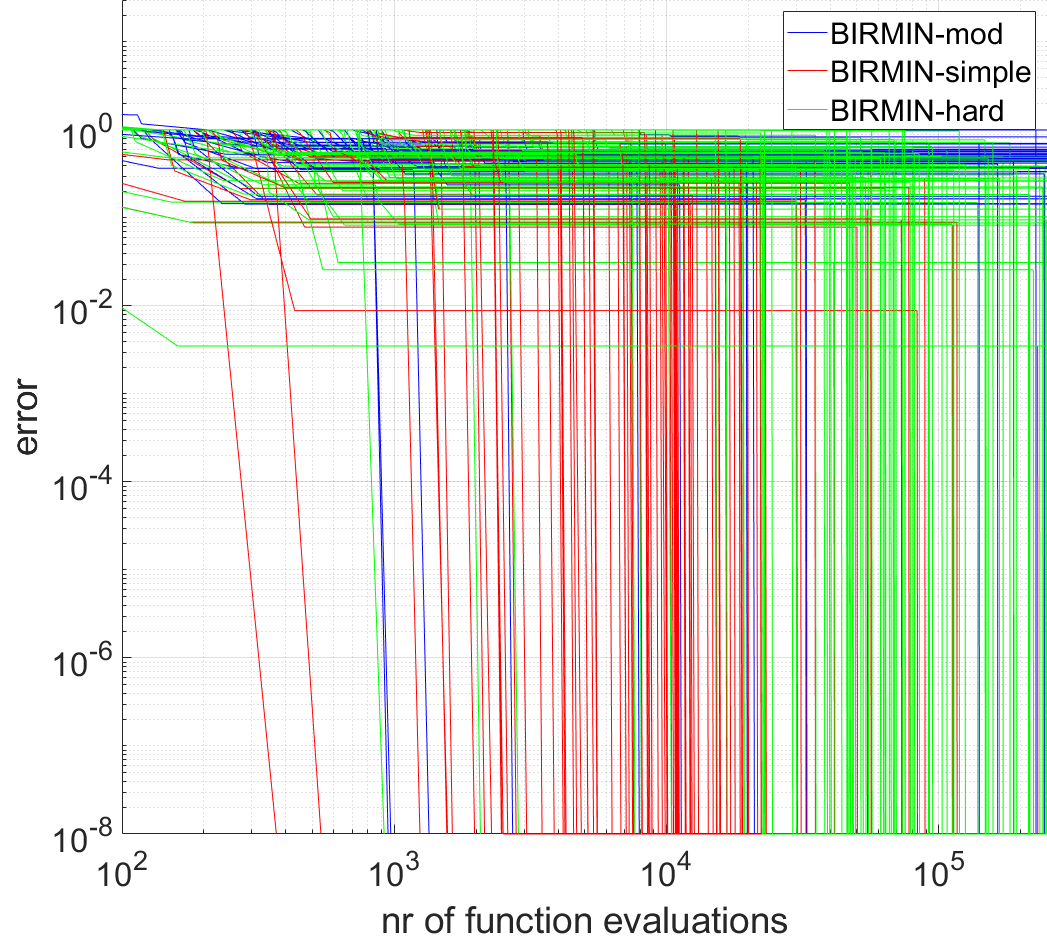}\\
        AGSK & LSHADE & BIRMIN
    \end{tabular}
    \caption{Convergence plots of the three methods in dimension $D = 5$.}
    \label{fig:conv5}
\end{figure*}

\begin{figure*}
    \centering
    \begin{tabular}{ccc}
       \includegraphics[width=0.3\linewidth]{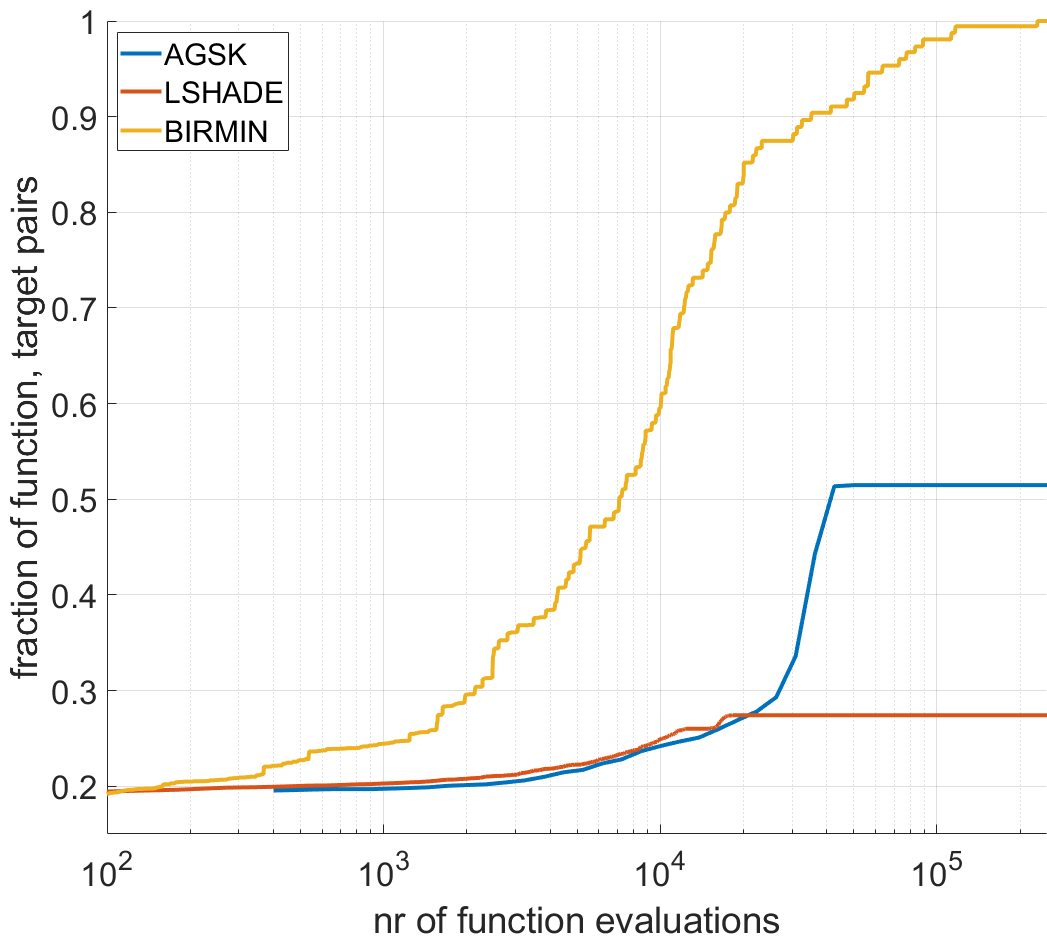}  &  \includegraphics[width=0.3\linewidth]{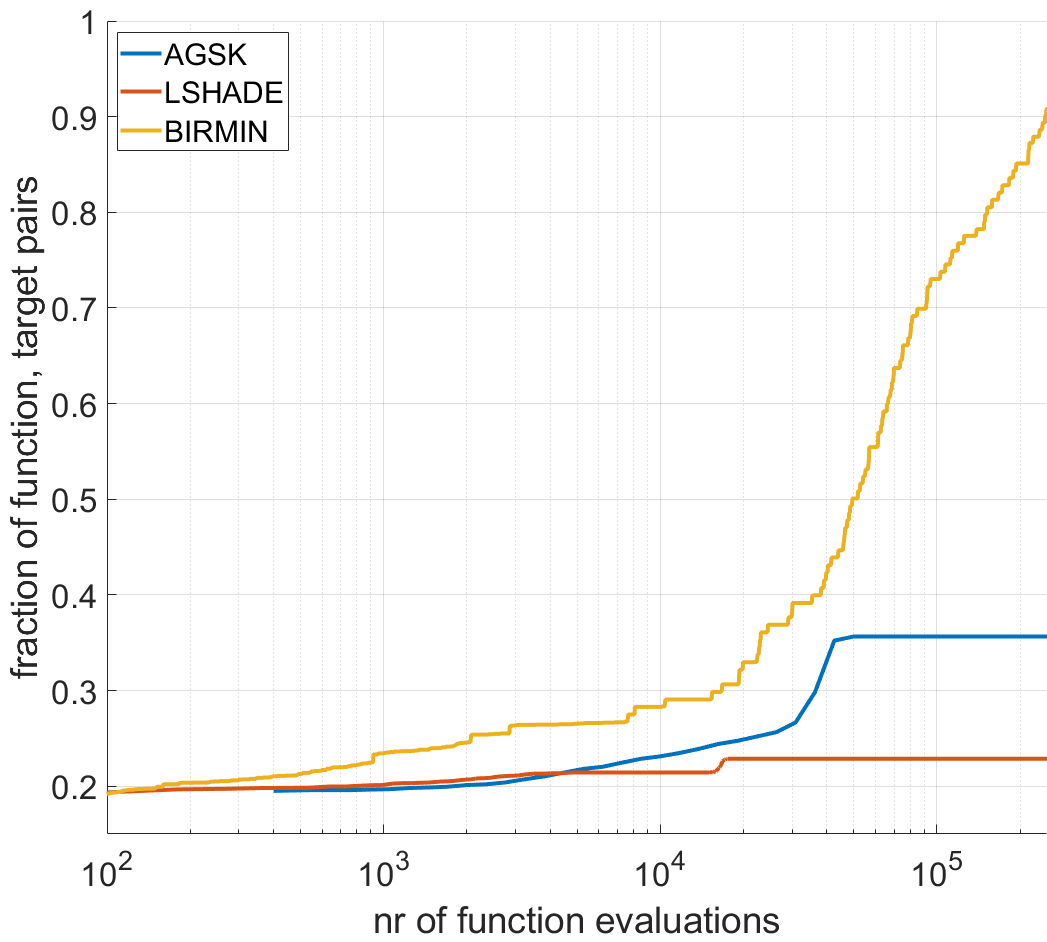} & \includegraphics[width=0.3\linewidth]{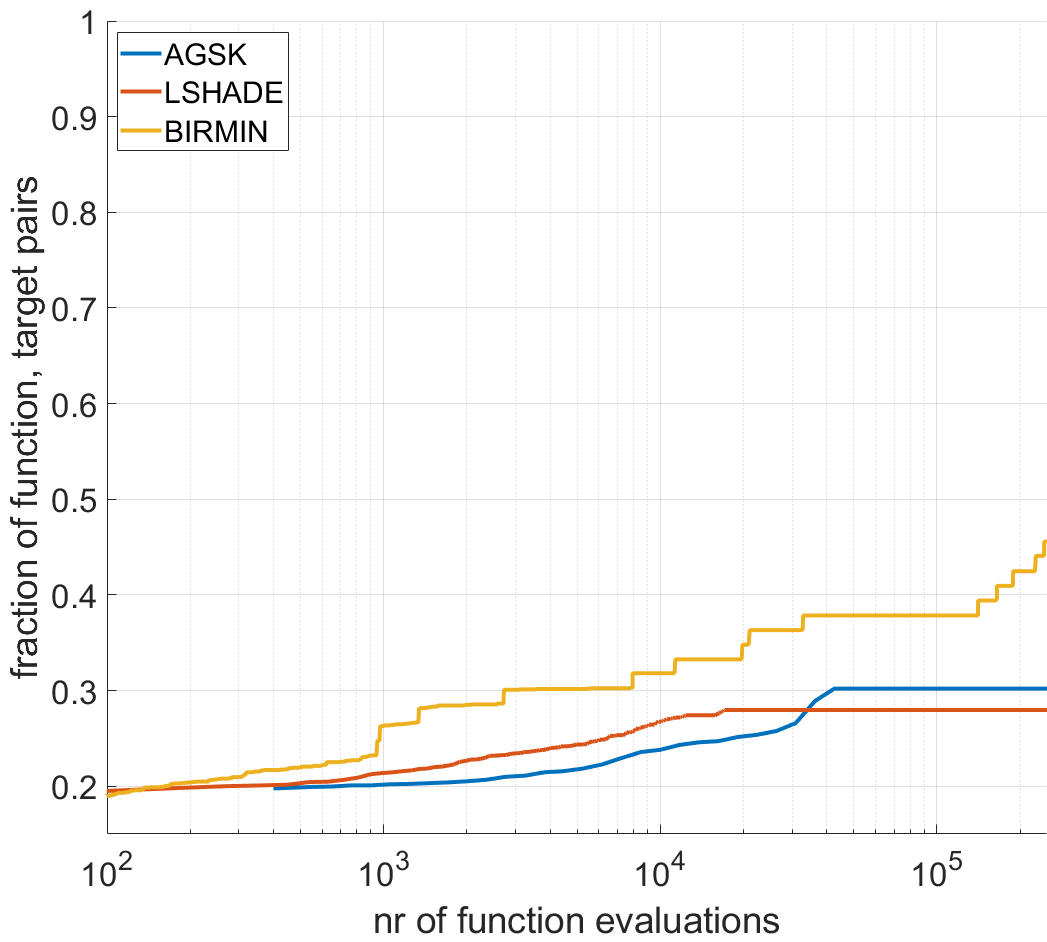}\\
        GKLS simple & GKLS hard & GKLS mod
    \end{tabular}
    \caption{Empirical cumulative distribution of simulated runtimes, measured in number of $f$-evaluations for the $51$  targets $10^{[-8..2]}$ in dimension $D = 5$.}
    \label{fig:frac5}
\end{figure*}

\section{Computational Analysis}
Although the classes in the same dimension might share some characteristics, what is arguably more important is whether it means that different algorithms will ``perform'' in the same way on these classes of problems. To study this effect, we set up the following computational analysis. We run three state-of-the-art methods both from the evolutionary computation (EC) and deterministic optimization communities on the ``canonical'' GKLS-generated problems in dimensions 5 and 10 (which uses the same ``simple'' and ``hard'' parameters as dimension 5). We also construct a new class (``mod'') 50 of GKLS-generated problems (again, in $D = 5$, and $10$) by the following procedure:
\begin{itemize}
    \item Each problem $i=1,\dots,50$ in this class will have different values of the parameters $\text{type}_i$, $d_i$, $r_i$, and $h_i$, but the same $f_i^* = -1$.
    \item $\text{type}_i$ is decided by a coin flip between types D and ND (with the same probability).
    \item $d_i$ is a uniformly distributed random number on [0,1].
    \item $r_i = d_i/u_i$, where $u_i$ is a uniformly distributed random integer on [2,10], i.e. $r_i \in [d_i/2, d_i/10]$.
    \item $h_i = \text{round}(10^{c_i})$, where $c_i$ is a uniformly distributed random number on [1,3], i.e. $h \in [10,10^3]$.
\end{itemize}

From the EC side, we chose two methods to run on the GKLS-generated problems. The first selected method was Adaptive Gaining-Sharing Knowledge (AGSK), which was the runner-up of the CEC’20 competition. The algorithm improved the original GSK algorithm by adding adaptive settings to its two control parameters: the knowledge factor and ratio, which control junior and senior gaining and sharing phases during the optimization process \cite{mohamed2020evaluating}. The second method is L-SHADE or Success-history based adaptive differential evolution with linear population size reduction \cite{tanabe2014improving}. This metaheuristic method has its basis in adaptive differential evolution, which involves success-history-based parameter adaptation. The proposed method then provides an extension in the form of using linear population size reduction, which results in population size reduction according to a linear function. From the deterministic methods, we selected the best-performing method from a recent extensive numerical study \cite{stripinis2022extensive}, which evaluated 64 derivative-free algorithms on the test problems from the DIRECTGOLib \cite{linas_stripinis_2022_6617799} and on the GKLS generator. The chosen method was BIRMIN \cite{paulavivcius2020globally}, which is a globally-biased hybridized version of the BIRECT method \cite{paulavivcius2018global}, which was also ``trained'' on the GKLS generator.

\begin{figure*}
    \centering
    \begin{tabular}{ccc}
       \includegraphics[width=0.3\linewidth]{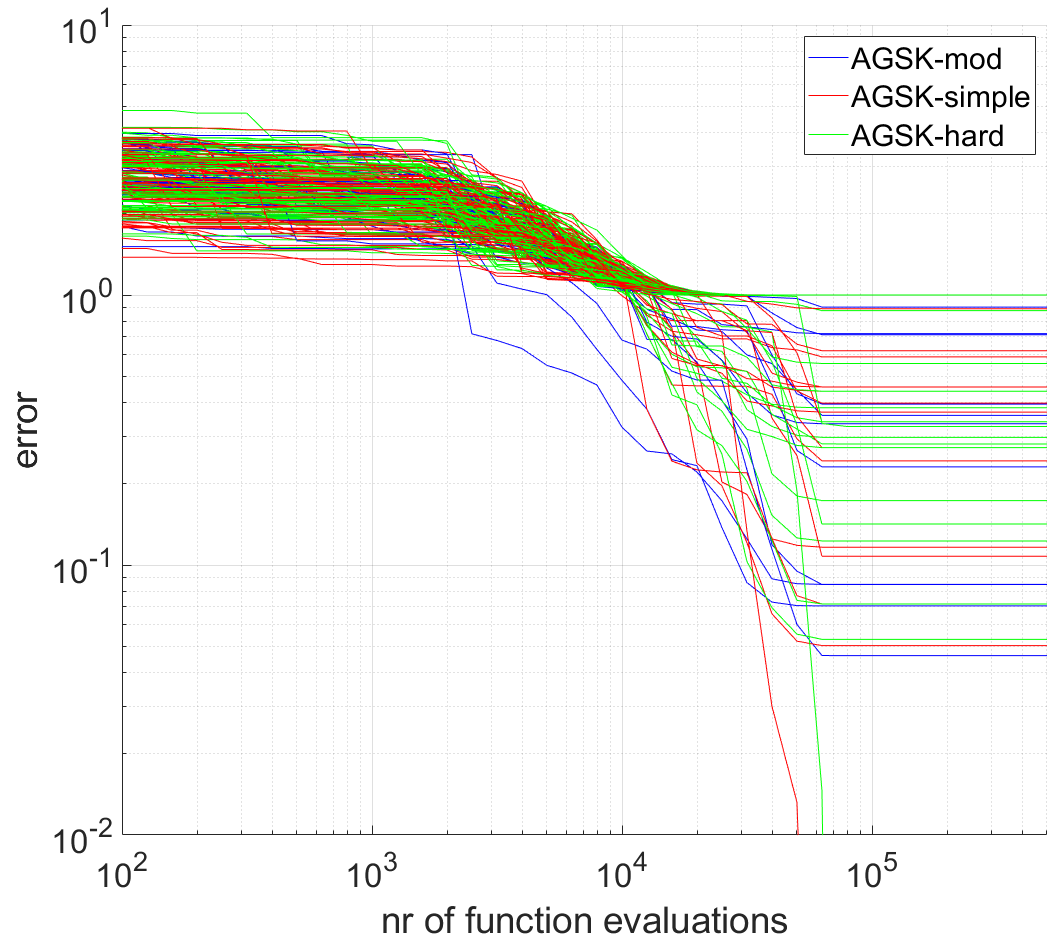}  &  \includegraphics[width=0.3\linewidth]{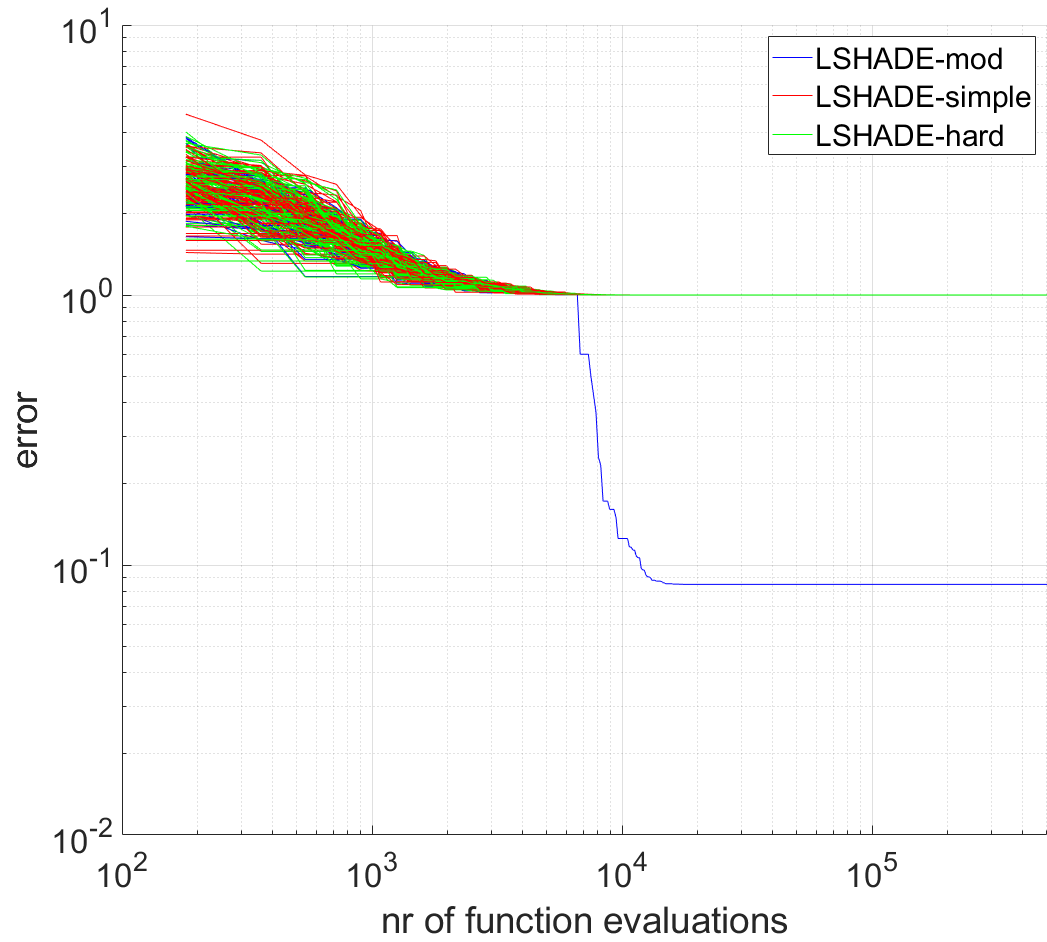} & \includegraphics[width=0.3\linewidth]{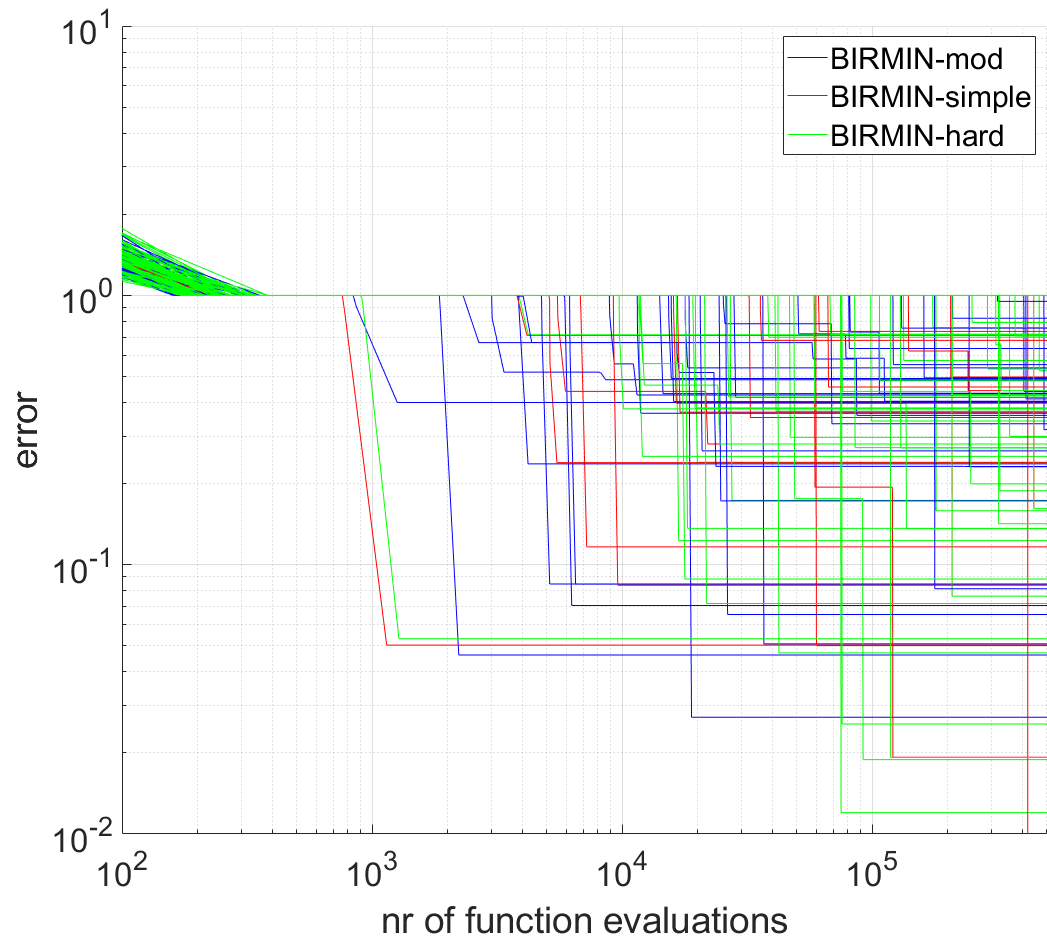}\\
        AGSK & LSHADE & BIRMIN
    \end{tabular}
    \caption{Convergence plots of the three methods in dimension $D = 10$.}
    \label{fig:conv10}
\end{figure*}

\begin{figure*}
    \centering
    \begin{tabular}{ccc}
       \includegraphics[width=0.3\linewidth]{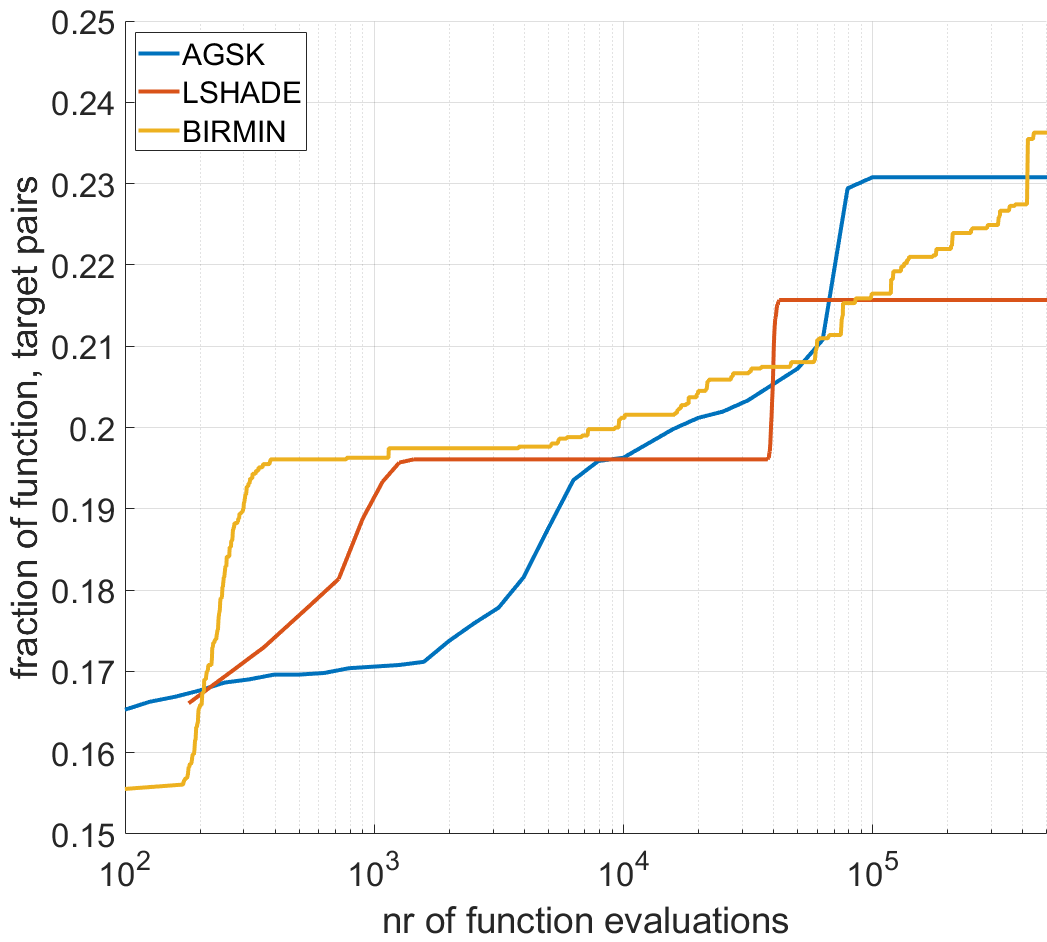}  &  \includegraphics[width=0.3\linewidth]{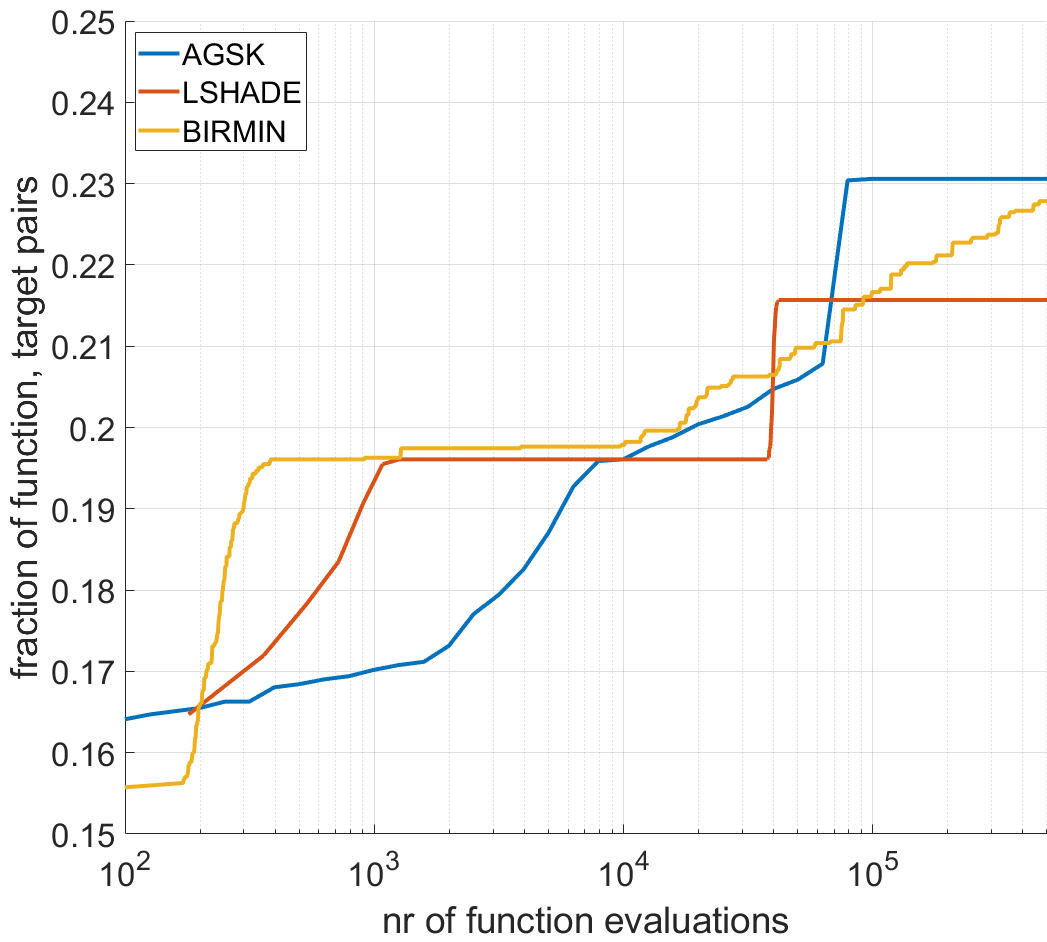} & \includegraphics[width=0.3\linewidth]{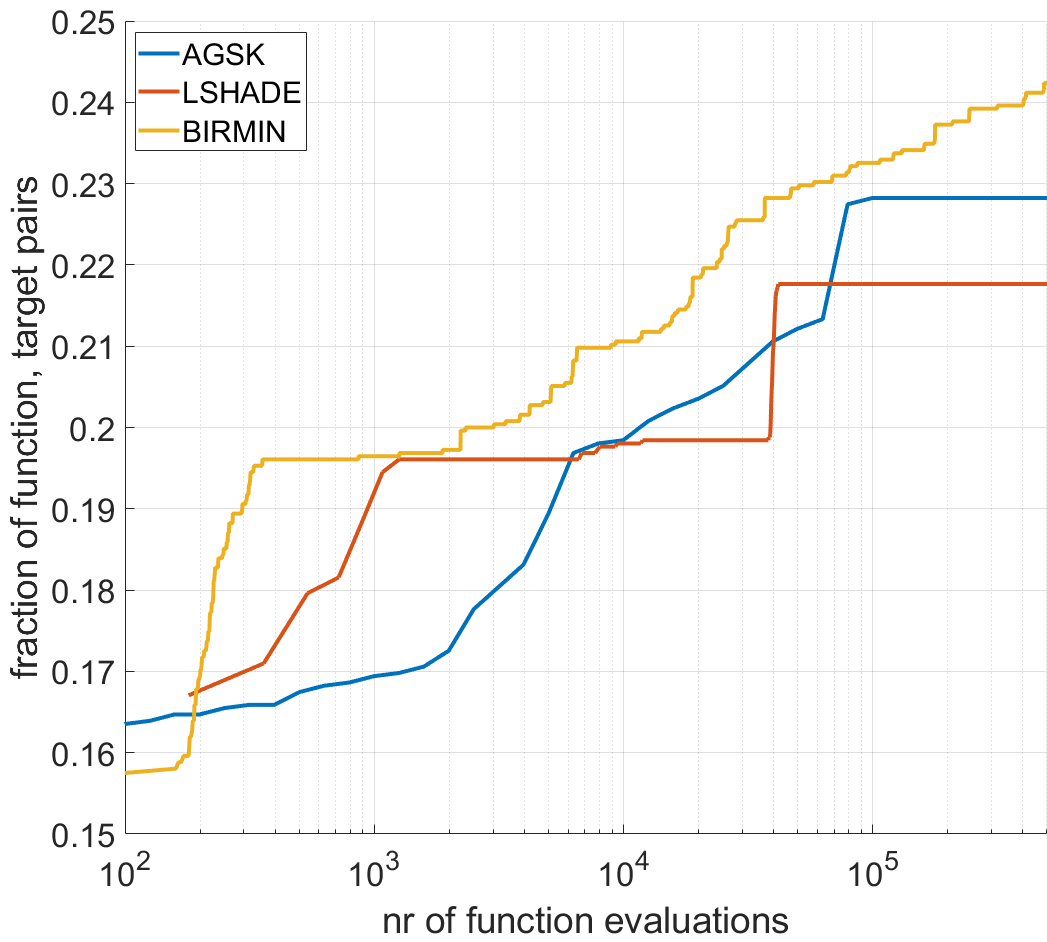}\\
        GKLS simple & GKLS hard & GKLS mod
    \end{tabular}
    \caption{Empirical cumulative distribution of simulated runtimes, measured in number of $f$-evaluations for the $51$  targets $10^{[-8..2]}$ in dimension $D = 10$.}
    \label{fig:frac10}
\end{figure*}

The implementation and parameter choices for LSHADE and AGSK were taken from \cite{biswas2021improving} (the implementations are available at the GitHub\footnote{https://github.com/subhodipbiswas/MadDE} of one of the authors). The implementation and parameter choices for BIRMIN were taken from the DIRECTGOLib \cite{linas_stripinis_2022_6617799}.

For the numerical comparison, we chose to run each of the three methods once on every problem from each of the three classes (100 problems in both ``simple'' and ``hard'' classes, and 50 problems in the ``mod'' class) in dimensions $D = [5,10]$, with a budget of $5\cdot 10^4 \cdot D$ available function evaluations.  For every run, if the objective function value of the resulting solution was less than or equal to 1E–8, it was considered as zero. All algorithms were run in a MATLAB R2022a, on a PC with a 3.2 GHz Core I5 CPU, 16 GB RAM, and Windows 10. The code for the experiments (and the generator for the test problems) can be found at the author's Github\footnote{https://github.com/JakubKudela89/GKLS-GECCO}. 

We begin the discussion on the results of the computations on classes in dimension $D = 5$. The convergence plots of the three methods on the three classes are shown in Figure \ref{fig:conv5}, while the Empirical cumulative distributions (ECDs) of simulated runtimes, measured in the number of function evaluations for 51 targets $10^{[-8..2]}$ (similar analysis which is done in the COCO platform) are shown in Figure \ref{fig:frac5}. The first thing one can find in these results is that although the ``simple'' and ``hard'' classes are composed of similar problems (as discussed above), there is a noticeable difference in the behavior of the three algorithms. On the ``simple'' class all three algorithms were able to find either a good or the optimal solutions faster than on the ``hard'' class. There is also a quite large difference between the performance of the three different methods - BIRMIN clearly dominated the two EC methods (and was able to find the optimal solution for a large portion of the problems), and AGSK turned out to be better at finding good solutions at the later stages of the search than LSHADE. However, both AGSK and LSHADE experienced a plateau in their convergence between $10^4$ and $4\cdot 10^4$ function evaluations and got stuck in local optima. These methods would undoubtedly benefit from using restart strategies \cite{tanabe2015tuning} or from hybridization with model-based optimization \cite{okulewicz2021benchmarking}, but we did not pursue these possibilities.

The results change quite dramatically when looking at the ``mod'' class. Although the performance of BIRMIN is still superior to that of the two EC methods, the margin narrowed substantially. What is more, the relative performance (against the ``hard'' class) of AGSK decreased, while for LSHADE it increased. For all three classes, the local minimum of the ``big'' paraboloid (error value 1) was found by every method within a few hundred function evaluations for BIRMIN and a few thousand function evaluations for the EC methods.

\begin{figure*}
    \centering
    \begin{tabular}{ccc}
       \includegraphics[width=0.3\linewidth]{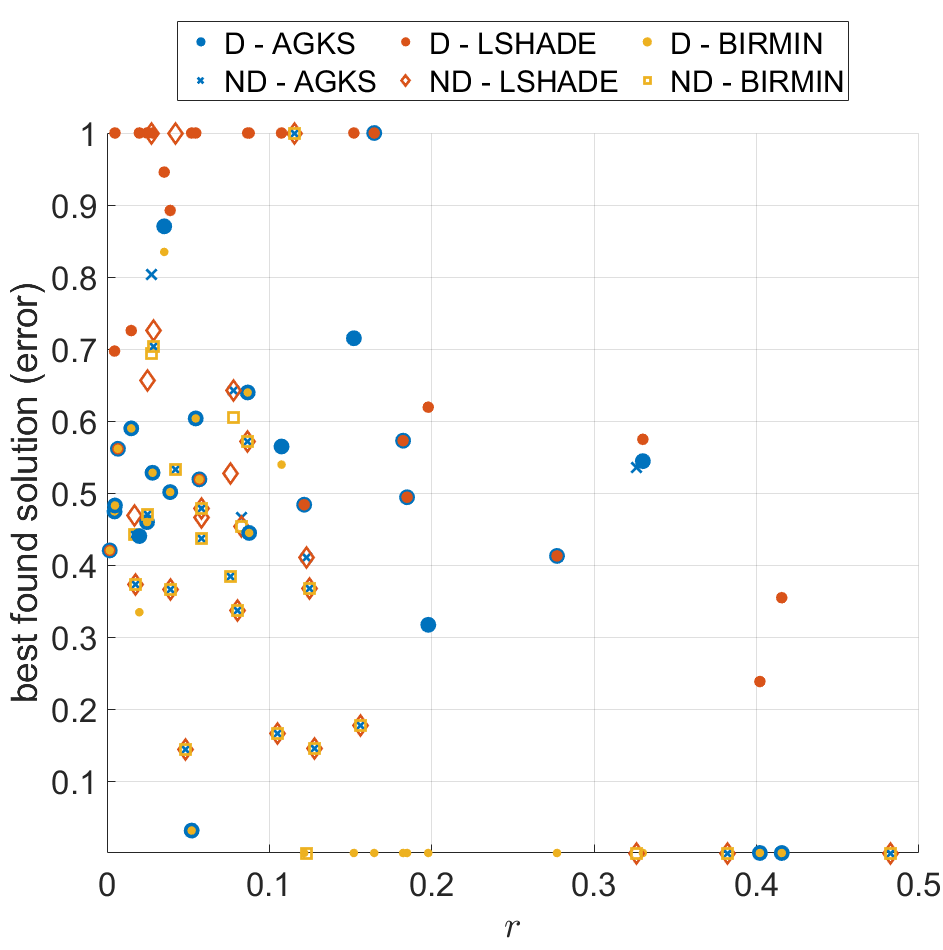}  &  \includegraphics[width=0.3\linewidth]{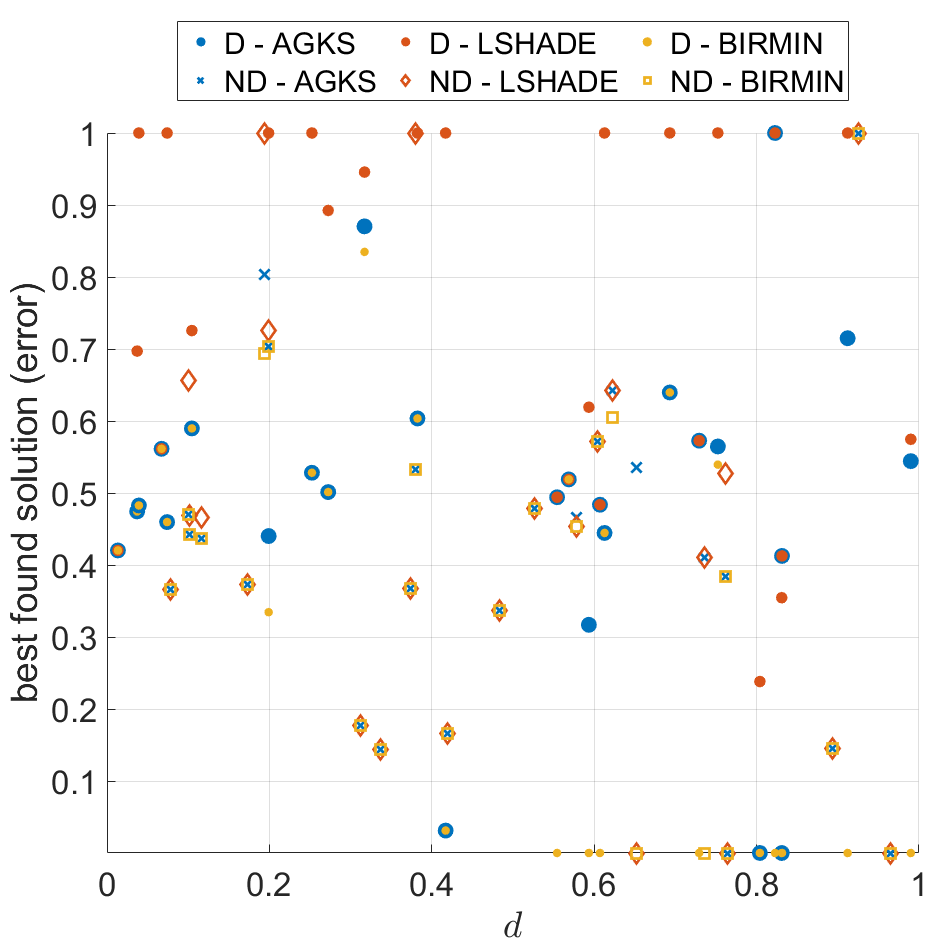} & \includegraphics[width=0.3\linewidth]{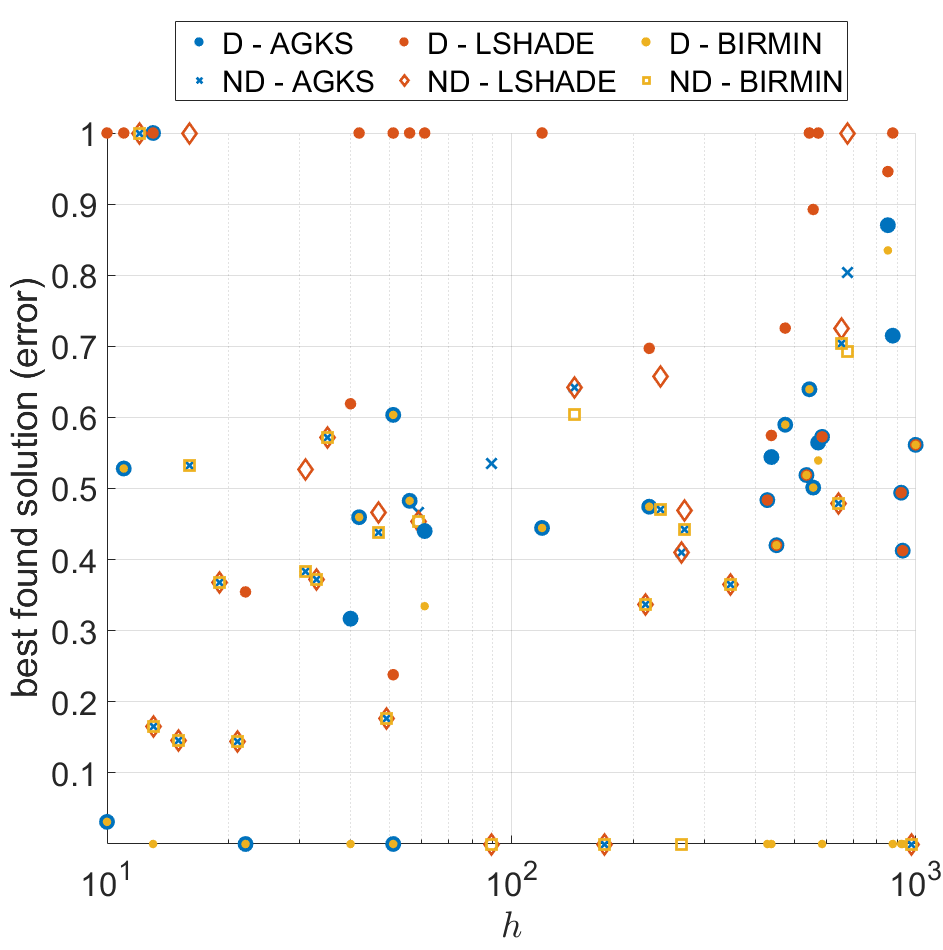}\\
    & dimension $D = 5$ \\[5mm]
    \includegraphics[width=0.3\linewidth]{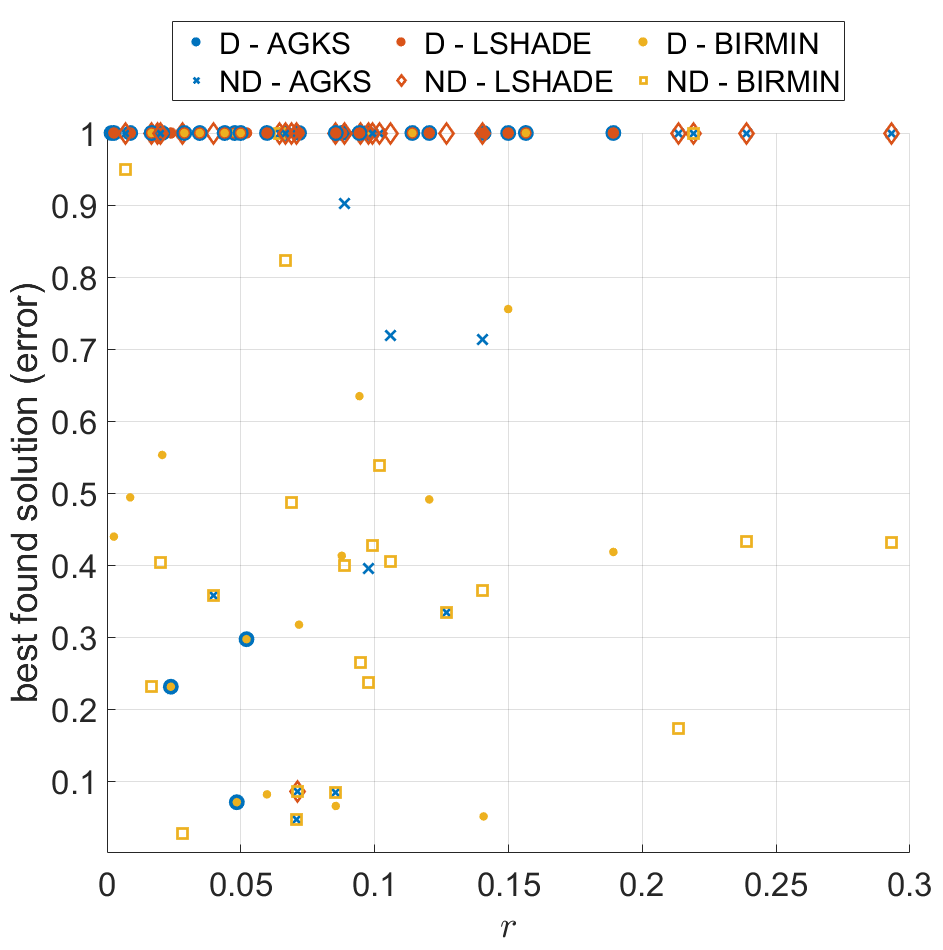}  &  \includegraphics[width=0.3\linewidth]{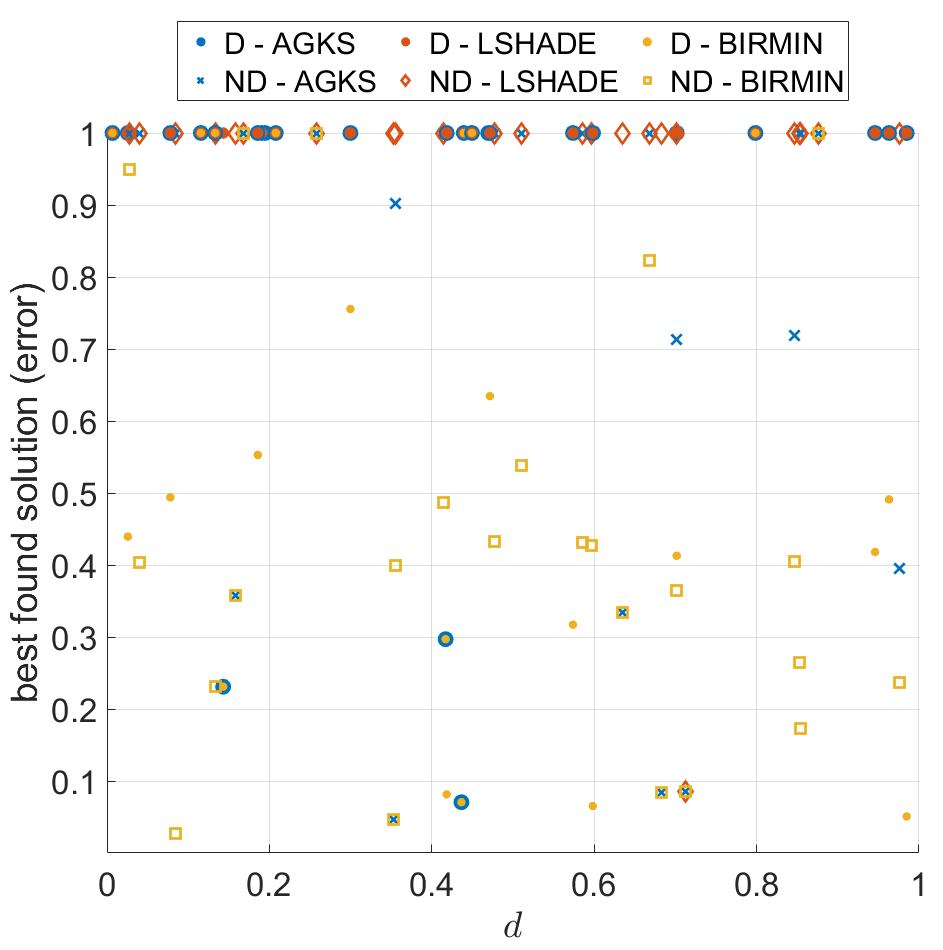} & \includegraphics[width=0.3\linewidth]{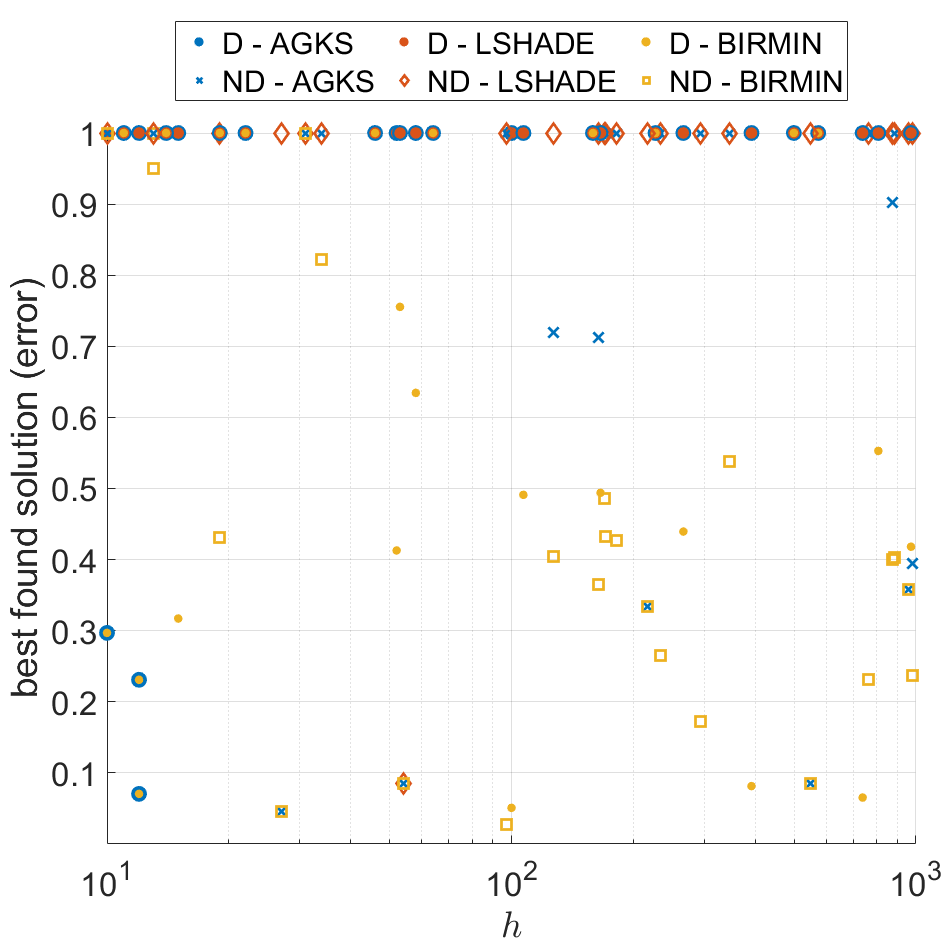}\\
    & dimension $D = 10$
    \end{tabular}
    \caption{Relashionship between $r, d, h$, type, and the best function value found on the ``mod'' class problems for the three methods for dimensions $D=5$ and $D = 10$.}
    \label{fig:paramplots}
\end{figure*}

The results for dimension $D = 10$, shown in Figures \ref{fig:conv10} and \ref{fig:frac10} show another substantial change. Looking at the ECD plot (notice the change in the range of the y-axis compared to a similar figure in dimension $D=5$) in Figure \ref{fig:frac10}, we can find that all three classes are basically equivalent. The plateaus in the ECD plots between the 0.19 and 0.20 values on the y-axis indicate the points where the methods found the local minimum of the ``big'' paraboloid. Although this happened relatively early for all methods, finding better local optima proved to be extremely challenging. And the global optimum was found only twice in the ``simple'' class (once by BIRMIN and once by AGSK on different problems) and once in the ``hard'' class (by AGSK). In the ``mod'' class, no method was able to find the global optimum for any of the problems, but there were more ``good'' local optima being found, especially by DIRMIN (hence the slightly better ECD plot).

As the problems ``mod'' class were generated using different values of the type, $r, d$, and $h$ parameters, we can study their effect on the solutions obtained by the three methods. This relationship is depicted in Figure \ref{fig:paramplots}. We focus first on dimension $D = 5$. First, we can see that the type of the problem (differentiable or non-differentiable) had practically no effect on the quality of the obtained solution. Rather intuitively, larger values of $r$ (region of attraction of the global minimizer) resulted in the methods finding the global optimum more easily. What is not so intuitive is the dependence on $d$ (distance from the global minimizer to the local minimum of the ``big'' paraboloid). Here, we can see that larger values of $d$ meant that the methods had a better chance of finding the global optimum. This was probably caused by the way $r$ was generated (as a fraction of $d$), which meant that problems with larger values of $d$ had also larger values of $r$. The last parameter we look at is $h$ (the number of local minima). While the difficulty of finding the global optimum seems to be unaffected by $h$, the parameter did have an effect if the best-found solution was not the global optimum. Increasing $h$ had the effect of worsening the best-found solution.

\begin{figure*}
    \centering
    \begin{tabular}{cc}
        \includegraphics[width = .35\linewidth]{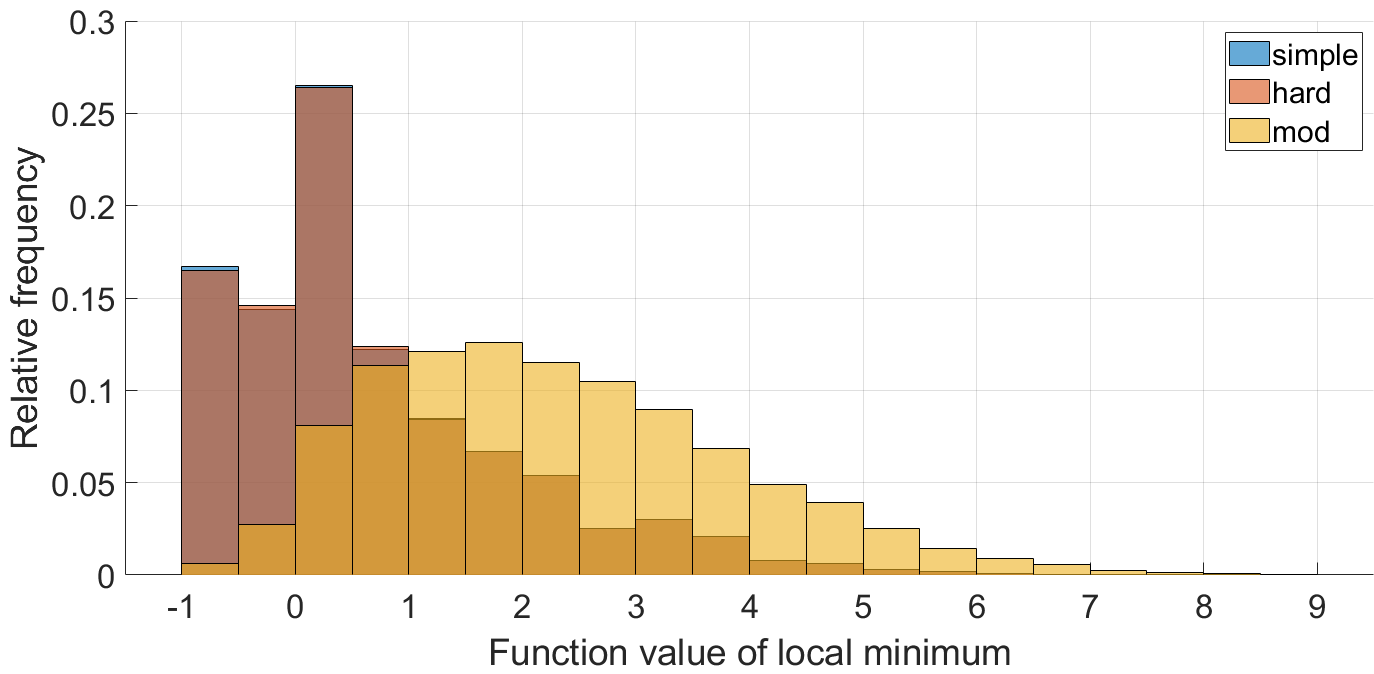} &  \includegraphics[width = .35\linewidth]{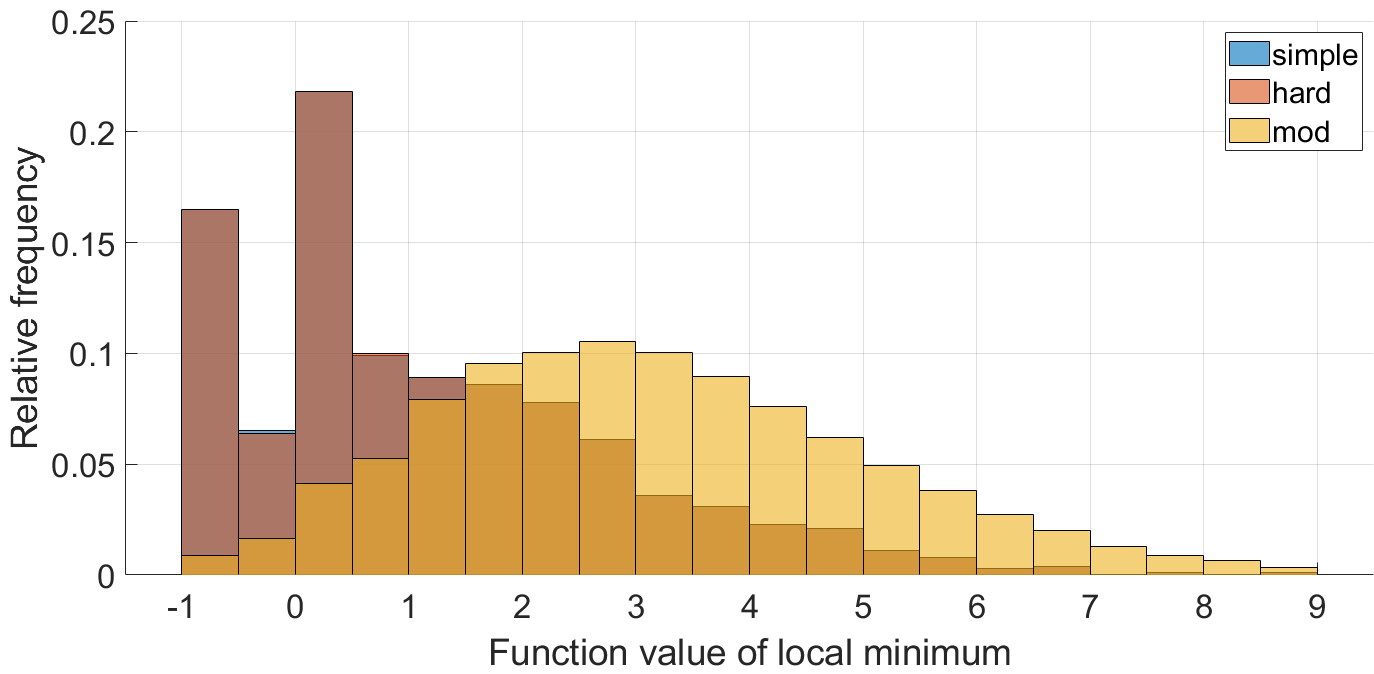} \\
        dimension $D = 5$ & dimension $D = 10$
    \end{tabular}
    \caption{Relative frequencies of function values of local minima of the different classes.}
    \label{fig:hist_local}
\end{figure*}

\begin{figure*}
    \centering
    \begin{tabular}{cc}
        \includegraphics[width = .35\linewidth]{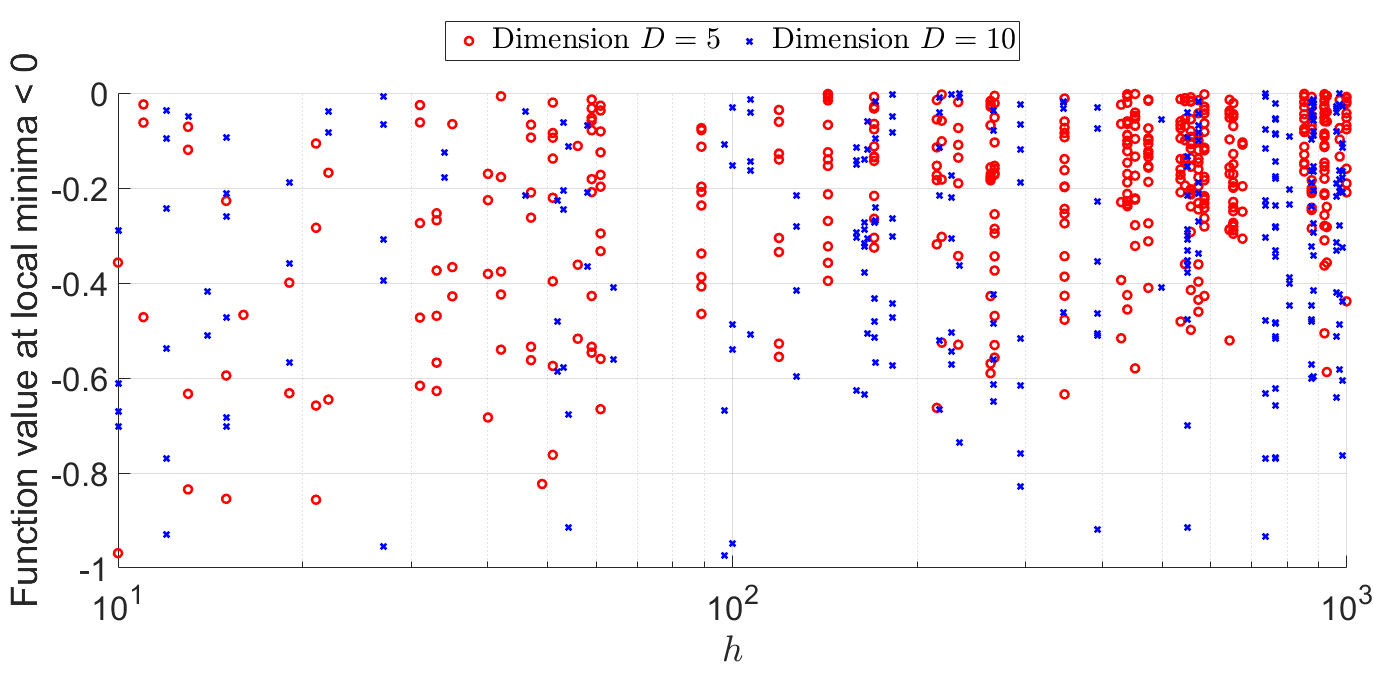} &  \includegraphics[width = .35\linewidth]{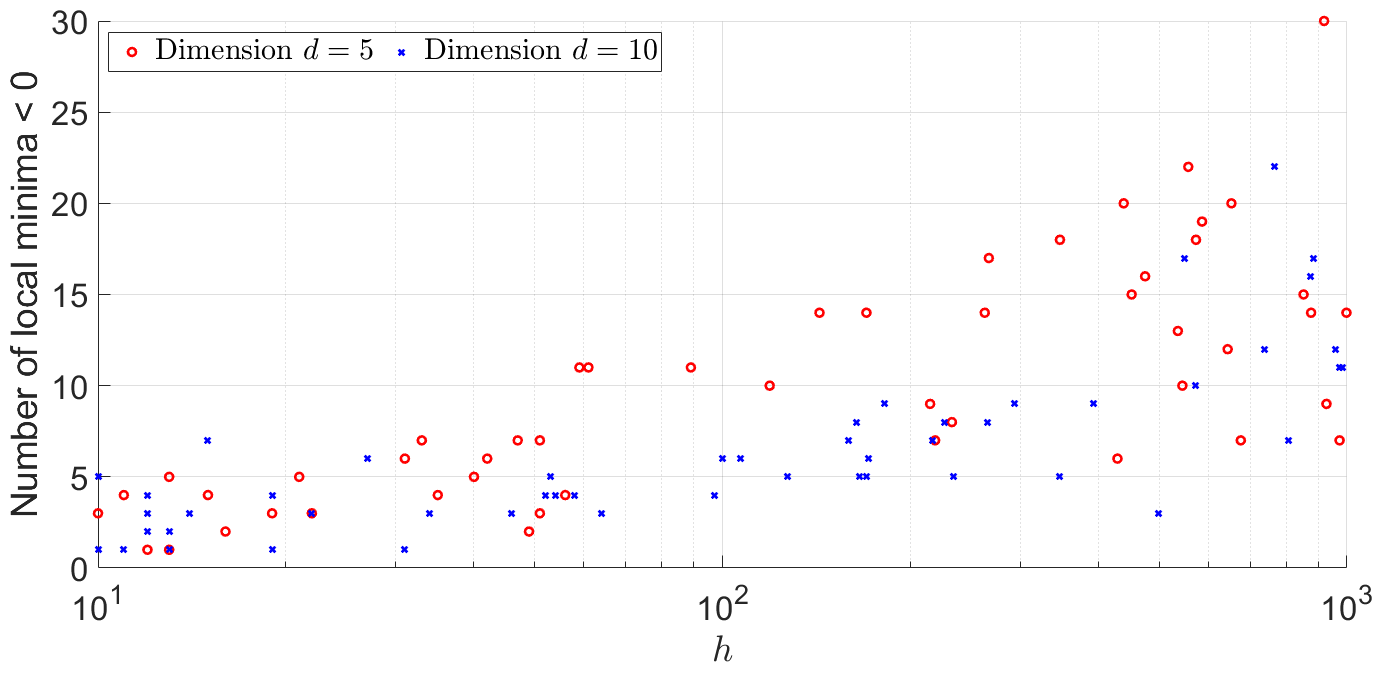} \\
        a) & b)
    \end{tabular}
    \caption{Scatter plots of the dependence of a) function values of local minima with function value < 0, b) number of local minima with function value < 0 on $h$ and dimension $D$.}
    \label{fig:scat}
\end{figure*}

We can trace this effect back to the function plots shown in Figure \ref{fig:f_plots}, where having more local minima meant that they became shallower. We can see this effect in Figure \ref{fig:hist_local}, where the relative frequencies of function values of the local minima for the three classes are plotted (the ``simple'' and ``hard'' classes have almost the same values of all the local minima - this is again, a consequence of the implemented pseudorandom number generator). 

In dimension $D=10$, the dependence of the best-found function value on the different values of the parameters vanishes. This is in line with the observation that the behavior of the three methods on three different classes in dimension $D = 10$ almost did not change (Figure \ref{fig:frac10}). However, one would still expect a similar dependence on $h$ as was seen in dimension $D = 5$. Especially since Figure \ref{fig:hist_local} shows a noticeable shift of the function values of the local minima to higher values for dimension $D = 10$ (when compared to dimension $D = 5$). This is explained in Figure \ref{fig:scat}, where we can see that although there are really fewer local minima in dimension $D = 10$ with negative function values, they are spread out more evenly than in the case of dimension $D = 5$, even when the number of local minima is high. 

The functions that the GKLS generator produces are of the ``needle in a haystack'' kind \cite{droste2002optimization}. If one does not stumble upon the region of the space where the global minimizer (or at least a local minimizer with ``good'' function value) has its region of attraction, either by chance (in the case of the EC methods) or by space partition (whose cost is bound to be exponential in the dimension $D$), the solution one gets is the local minimum of the ``big'' paraboloid. The functions give no hints on where such good points might be.

\section{Exploratory Landscape Analysis}
We use ELA features to show how the GKLS-generated problems compare to the BBOB and CEC 2014 benchmark suits. In order to calculate the ELA features, we used the flacco library \cite{kerschke2019comprehensive}. We chose ELA feature sets which only require samples of input and function value pairs: {\tt ela\_distr}, {\tt ela\_meta}, {\tt disp}, {\tt nbc}, {\tt pca}, and {\tt ic}, and dimension $D = 10$ for all considered suits. It was recently shown that the ELA features are sensitive to sampling strategy \cite{renau2020exploratory} and function transformations \cite{vskvorc2022comprehensive,vskvorc2021effect}. We chose to ignore the features that were sensitive to function transformations and used uniform sampling with $250\cdot D$ samples for the computation of the ELA features in all suits. There were 24 problems in the BBOB suit, 30 problems in the CEC 2014 suit, 100 problems in the GKLS ``simple'' and ``hard'' classes, and 50 problems in the GKLS ``mod'' class.

We then followed the methodology described in \cite{vskvorc2020understanding} for the selection and visualization of the relevant ELA features. The features that produced constant results on every problem and those that produced invalid values were removed. Another batch of features that got removed were the highly correlated ones. The 14 features that remained, along with their maximum and minimum values on the different suits are shown in Table \ref{t:ela}. We can see that the ranges of values of the chosen features for the three GKLS-generated suits are very similar. The ranges of the feature values of the GKLS-generated suits are also generally narrower than that of both BBOB and CEC 2014 suits. Both of these observations should be expected, as (as we have seen in the previous section) the GKLS generator produces somehow limited types of functions. One interesting observation can be made regarding the value of the {\tt ela\_distr.number\_of\_peaks} on the GKLS ``mod'' suit. Although in this suit there are problems with hundreds of local minima, they are practically all too shallow to be noticeable in the sampled points.

\begin{table*}[]
\caption{Minimum and Maximum values of the relevant ELA features on the different benchmark sets. Extremal values are highlighted in bold.}
\label{t:ela}
\resizebox{1\linewidth}{!}{
\begin{tabular}{l|rr|rr|rr|rr|rr}
                                 & \multicolumn{2}{c|}{BBOB} &   \multicolumn{2}{c|}{CEC 2014}    & \multicolumn{2}{c|}{GKLS simple} & \multicolumn{2}{c|}{GKLS hard}  & \multicolumn{2}{c}{GKLS mod}  \\
                                 & \multicolumn{1}{c}{min}  & \multicolumn{1}{c|}{max} & \multicolumn{1}{c}{min}      & \multicolumn{1}{c|}{max} & \multicolumn{1}{c}{min}         & \multicolumn{1}{c|}{max} & \multicolumn{1}{c}{min}       & \multicolumn{1}{c|}{max} & \multicolumn{1}{c}{min}      & \multicolumn{1}{c}{max} \\ \hline
{\tt ela\_distr.skewness}              & \textbf{-2.97E+00}       & \textbf{8.28E+00}       & -6.63E-01                    & 6.47E+00                & 1.02E-01                        & 4.18E-01                & 1.89E-01                      & 4.22E-01                & 2.01E-01                     & 4.46E-01                \\
{\tt ela\_distr.kurtosis}              & \textbf{-4.94E-01}       & \textbf{9.67E+01}       & -3.38E-01                    & 6.50E+01                & -3.53E-01                       & 2.15E-01                & -3.40E-01                     & 1.66E-01                & -3.56E-01                    & 2.47E-01                \\
{\tt ela\_distr.number\_of\_peaks}       & \textbf{1.00E+00}        & 1.80E+01                & \textbf{1.00E+00}            & \textbf{2.60E+01}       & \textbf{1.00E+00}               & 2.00E+00                & \textbf{1.00E+00}             & 2.00E+00                & \textbf{1.00E+00}            & 2.00E+00                \\
{\tt ela\_meta.lin\_simple.intercept}   & \textbf{-9.17E+02}       & 9.62E+08                & 5.22E+02                     & \textbf{5.63E+10}       & 4.88E+00                        & 9.00E+00                & 4.80E+00                      & 8.92E+00                & 5.06E+00                     & 9.00E+00                \\
{\tt ela\_meta.lin\_w\_interact.adj\_r2}  & 2.14E-04                 & \textbf{1.00E+00}       & \textbf{-9.41E-04}           & 9.04E-01                & 6.91E-01                        & 8.98E-01                & 6.82E-01                      & 8.93E-01                & 7.04E-01                     & 8.95E-01                \\
{\tt ela\_meta.quad\_simple.adj\_r2}     & 3.98E-03                 & \textbf{1.00E+00}       & \textbf{-3.61E-03}           & 9.88E-01                & 9.93E-01                        & \textbf{1.00E+00}       & 9.94E-01                      & \textbf{1.00E+00}       & 9.90E-01                     & \textbf{1.00E+00}       \\
{\tt ela\_meta.quad\_w\_interact.adj\_r2} & 3.67E-05                 & \textbf{1.00E+00}       & \textbf{-1.26E-02}           & \textbf{1.00E+00}       & 9.94E-01                        & \textbf{1.00E+00}       & 9.95E-01                      & \textbf{1.00E+00}       & 9.91E-01                     & \textbf{1.00E+00}       \\
{\tt disp.ratio\_mean\_25}              & 8.43E-01                 & 1.00E+00                & 8.60E-01                     & \textbf{1.01E+00}       & 8.44E-01                        & 8.84E-01                & \textbf{8.37E-01}             & 8.84E-01                & 8.43E-01                     & 8.74E-01                \\
{\tt disp.ratio\_median\_02}            & 6.48E-01                 & \textbf{1.06E+00}       & 6.57E-01                     & 1.02E+00                & \textbf{6.21E-01}               & 7.22E-01                & 6.25E-01                      & 7.19E-01                & 6.32E-01                     & 7.20E-01                \\
{\tt nbc.nb\_fitness.cor}              & \textbf{-6.41E-01}       & \textbf{-1.78E-01}      & -6.30E-01                    & -1.90E-01               & -4.08E-01                       & -3.52E-01               & -4.20E-01                     & -3.58E-01               & -4.12E-01                    & -3.65E-01               \\
{\tt pca.expl\_var.cov\_init}           & \textbf{9.09E-02}        & \textbf{9.09E-01}       & 9.09E-02                     & \textbf{9.09E-01}       & 6.36E-01                        & 8.18E-01                & 6.36E-01                      & 8.18E-01                & 6.36E-01                     & 8.18E-01                \\
{\tt pca.expl\_var.cor\_init}           & \textbf{8.18E-01}        & \textbf{9.09E-01}       & \textbf{8.18E-01}            & \textbf{9.09E-01}       & \textbf{8.18E-01}               & 8.18E-01                & \textbf{8.18E-01}             & 8.18E-01                & \textbf{8.18E-01}            & 8.18E-01                \\
{\tt pca.expl\_var\_PC1.cov\_init}       & \textbf{1.07E-01}        & \textbf{1.00E+00}       & 1.10E-01                     & \textbf{1.00E+00}       & 4.97E-01                        & 7.48E-01                & 4.98E-01                      & 7.41E-01                & 5.23E-01                     & 7.52E-01                \\
{\tt pca.expl\_var\_PC1.cor\_init}       & 1.02E-01                 & 1.83E-01                & \textbf{9.94E-02}            & 1.76E-01                & 1.66E-01                        & \textbf{1.85E-01}       & 1.66E-01                      & 1.82E-01                & 1.64E-01                     & 1.80E-01               
\end{tabular}
}
\end{table*}

\begin{figure}
    \centering
    \includegraphics[width = \linewidth]{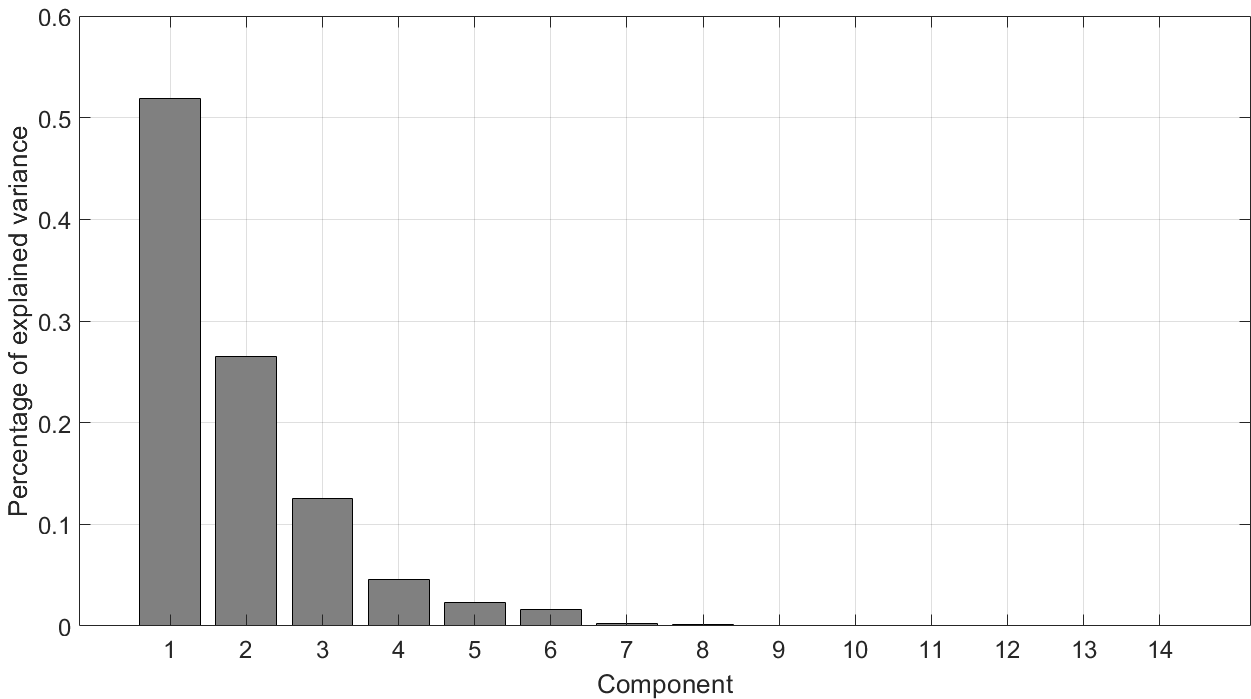}
    \caption{The amount of explained variance per component when performing PCA on the ELA features calculated on the combined set of all benchmark problems.}
    \label{fig:ela_hist}
\end{figure}

\begin{figure}
    \centering
    \includegraphics[width = \linewidth]{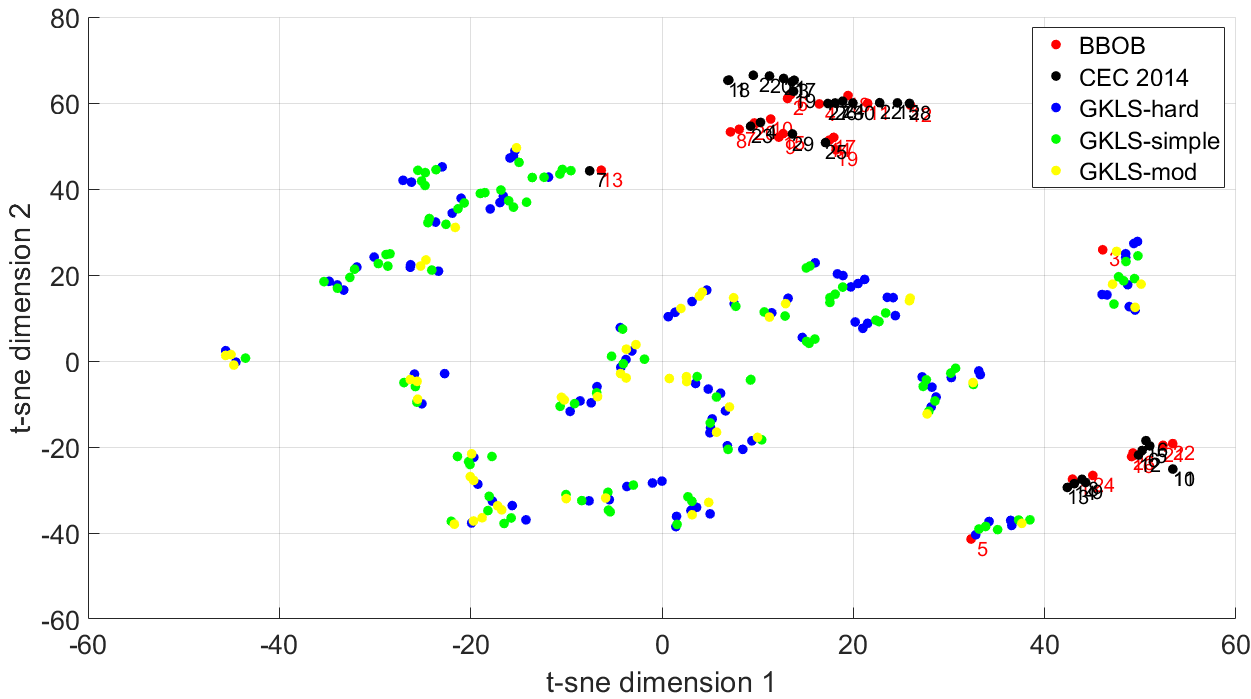}
    \caption{The t-sne visualization of the ELA features (after normalization and using the first seven components from the PCA) of the  benchmark sets.}
    \label{fig:ela_sne}
\end{figure}

For further analysis, the values of the ELA features on the different benchmark sets were normalized, and we used Principal Component Analysis (PCA) to reduce the number of features even further. Figure \ref{fig:ela_hist} shows a representation of the 14 PCA components obtained when comparing the ELA features (normalized) calculated on the combined set of CEC 2014, BBOB, and all GKLS-generated problems. Using the first 7 components explained 99.68\% of the variance. For visualizing the results, we used the t-Distributed Stochastic Neighbor Embedding (t-sne). In this visualization, which is shown in Figure \ref{fig:ela_sne}, benchmark problems that have similar ELA features should be shown close to each other. 

\begin{figure*}
    \centering
    \begin{tabular}{ccc}
       \includegraphics[width=0.3\linewidth]{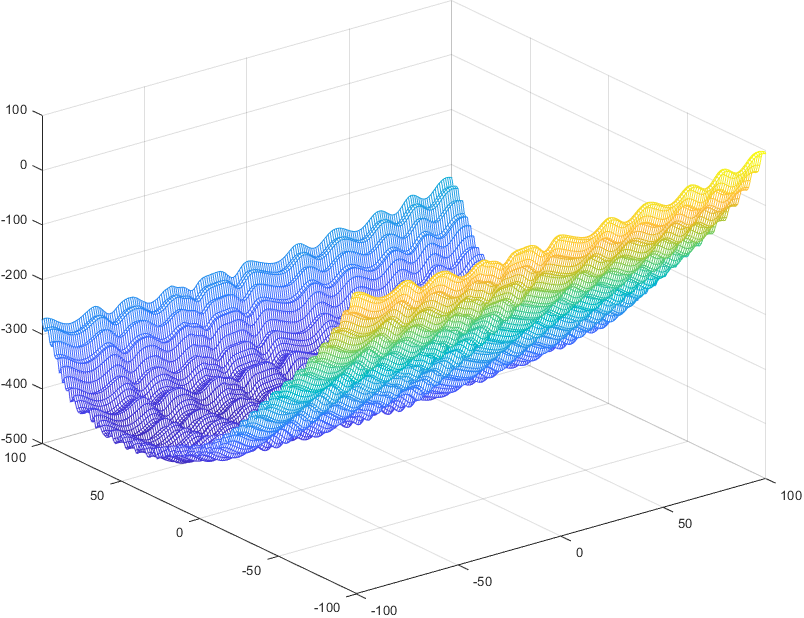}  &  \includegraphics[width=0.3\linewidth]{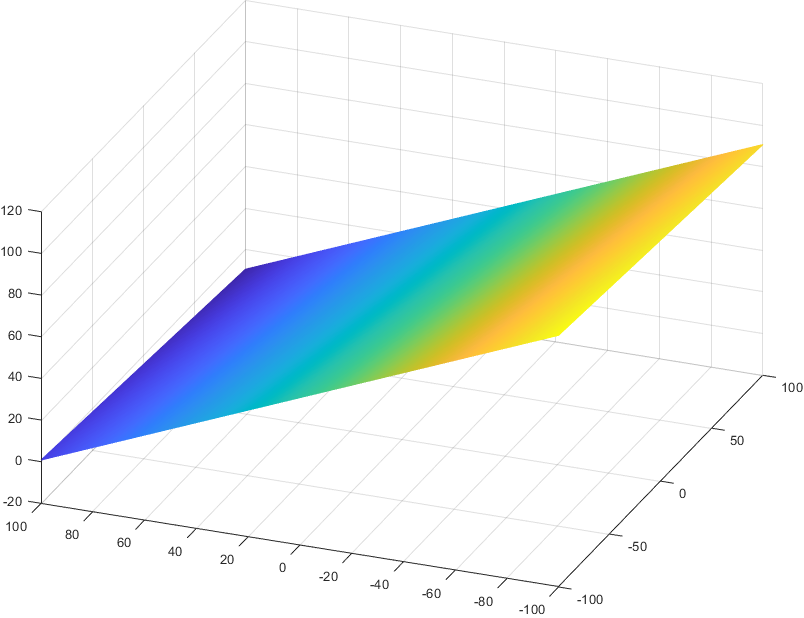} & \includegraphics[width=0.3\linewidth]{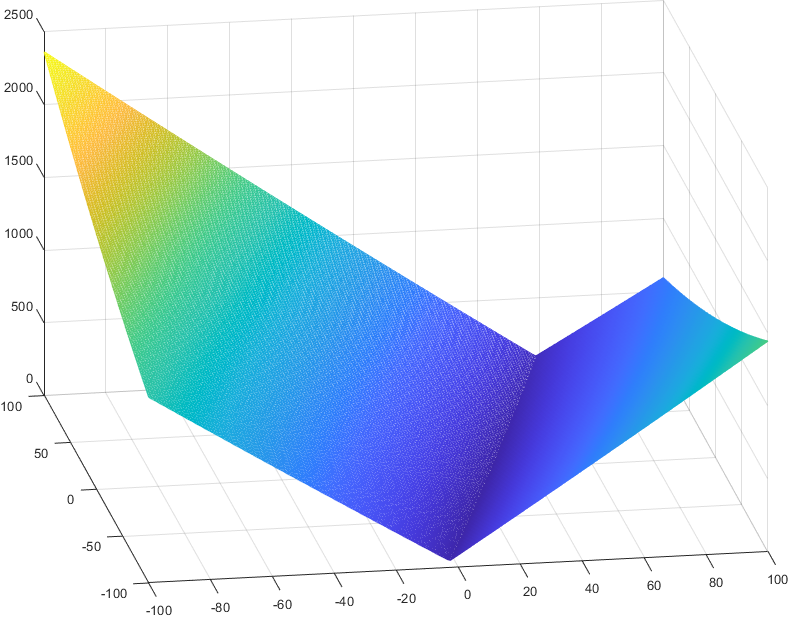}\\
        $f_3$ (Rastrigin separable) & $f_5$ (Linear slope) & $f_{13}$ (Sharp ridge)
    \end{tabular}
    \caption{The three functions from the BBOB suit that are the ``closest'' to some GKLS-generated instances.}
    \label{fig:cec}
\end{figure*}

We can see that the t-sne visualization grouped most of the functions from the BBOB and CEC 2014 suits together (in a few groups), while the ``similar'' problems generated in the three GKLS suites take up most of the space. It should be pointed out that the t-sne procedure is random, producing different plots every time. However, in our experiments, all plots looked qualitatively the same as Figure \ref{fig:ela_sne}. Interestingly, some of the GKLS-generated problems show common traits to a few problems in the other two benchmark suits. The three BBOB functions that are close to the GKLS-generated ones ($f_3$, $f_5$, and $f_{13}$) are shown (in dimension $D = 2$) in Figure \ref{fig:cec}. All three are probably close to some instance of the GKLS-generated ``big'' paraboloid, which would explain the closeness of the problems in the t-sne plot.

\section{Conclusion}
In this paper, we analyzed the problems constructed by the GKLS generator. In the computational analysis, we have shown that it produces what are basically  ``needle in a haystack '' problems which get extremely difficult to optimize as the problem dimension grows. The GKLS generator could be successfully used for benchmarking state-of-the-art methods in lower dimensions ($D = 5$) on some of the simpler instances. However, in the higher dimension ($D=10$), the performance of the three considered methods was hard to differentiate as the problems became extremely difficult for the given computational budget. This difficulty of the generated instances was also largely unaffected by the choice of parameters that the generator has. Although increasing the computational budget might bring additional insight, different restart strategies for the EC methods would have to be used, as they both plateaued long before reaching the maximum available function evaluations.

The type of function generated by GKLS (differentiable or non-differentiable) had practically no effect on its ``usefulness'' in benchmarking. It is possible that the GKLS generator could be modified to have a much ``deeper'' local minima. As the task of finding the global minimum is practically impossible in higher dimensions, having problems with lots of ``good'' local minima (i.e., better ones than the local minimum of the ``big'' paraboloid) could be useful for analyzing the exploration capabilities of optimization methods.
 
It is not very clear how one could meaningfully use the results of the computations of the ELA features or the t-sne plots on such ``needle in a haystack'' problems. The superficial closeness the some of the GKLS-generated problems to the three BBOB problems, and the large variation in the t-sne dimensions of the GKLS-generated problems hint at the limits where ELA can be meaningfully used. It is probably impossible to have any sample-based features that would both uncover that the problem is a ``needle in a haystack'' and be computationally tractable (as it would amount to finding the ``needle'' in a reasonable amount of function evaluations).

Lastly, it would be interesting to find real-world black-box continuous problems which have the ``needle in a haystack'' character and show if the methods that were ``trained'' on such problems (such as BIRMIN) really offer an advantage (at least in lower dimensions).

\begin{acks}
This work was supported by the IGA BUT No. FSI-S-23-8394 ``Artificial intelligence methods in engineering tasks''.
\end{acks}

\bibliographystyle{ACM-Reference-Format}
\bibliography{sample-base}


\end{document}